\documentclass[runningheads]{llncs}

\usepackage{eccv}

\usepackage{eccvabbrv}
\usepackage{listings}

\usepackage{graphicx}
\usepackage{booktabs}

\usepackage{orcidlink}

\begin{document}

\title{DETRPose: Real-Time End-to-End Multi-Person Pose Estimation via Modified Transformer Decoder and Novel Denoising Keypoints}

\titlerunning{DETRPose: Real-time DETR for Multi-Person Pose Estimation}

\author{Sebastian Janampa\inst{1}\orcidlink{0000-0001-9637-7408} \and
Marios Pattichis\inst{1}\orcidlink{0000-0002-1574-1827}}

\authorrunning{S. Janampa and M. Pattichis}

\institute{The University of New Mexico}

\maketitle

\begin{abstract}
Multi-person pose estimation (MPPE), which involves detecting body joint positions (keypoints) for every person in an image, is a fundamental task in computer vision. Despite recent advances, no transformer-based model currently achieves real-time performance. This work addresses the latency challenge by introducing DETRPose, the first family of real-time, end-to-end transformer models for multi-person 2D pose estimation. DETRPose significantly enhances the GroupPose decoder, enabling real-time inference. For training, a novel denoising keypoint technique is proposed to accelerate convergence. The varifocal loss is also extended for keypoints, termed Keypoint Similarity VariFocal loss, to improve query quality. Extensive evaluation demonstrates that DETRPose models achieve accuracy comparable to or exceeding that of leading alternatives while requiring five to ten times fewer training epochs. DETRPose-S matches the accuracy of YOLOv8-Pose-X and YOLO11-Pose-X on the COCO dataset (67.0 vs 67.3 and 67.2 in AP) with 81\% fewer parameters (11.5M vs 69.4M and 58.8M) and 52\% faster inference speed (2.39ms vs 5.23ms and 4.93ms). On the CrowdPose dataset, DETRPose-X has $13.1\times$ fewer FLOPs (232.3G vs 3048.1G) and only $2\%$ fewer precision (75.1 vs 76.6 in AP) than ED-Pose-SwinL-5S. On the OCHuman dataset, DETRPose-S surpasses all previous models, showing the robustness of DETRPose on out-of-distribution datasets. Code is available at \url{https://github.com/SebastianJanampa/DETRPose}
\end{abstract}    
\section{Introduction}

2D multi-person pose estimation (MPPE) is a critical preprocessing step for tasks including activity recognition, 2D-to-3D human pose estimation, and virtual reality. As a result, low latency is essential for real-time deployment. Although numerous studies on transformer models for MPPE focus on accuracy improvements, their high latency limits their applicability in real-time scenarios.

\begin{figure*}[!t]
	\centering
	
	\begin{tabular}{cc}
		\includegraphics[width=0.5\textwidth]{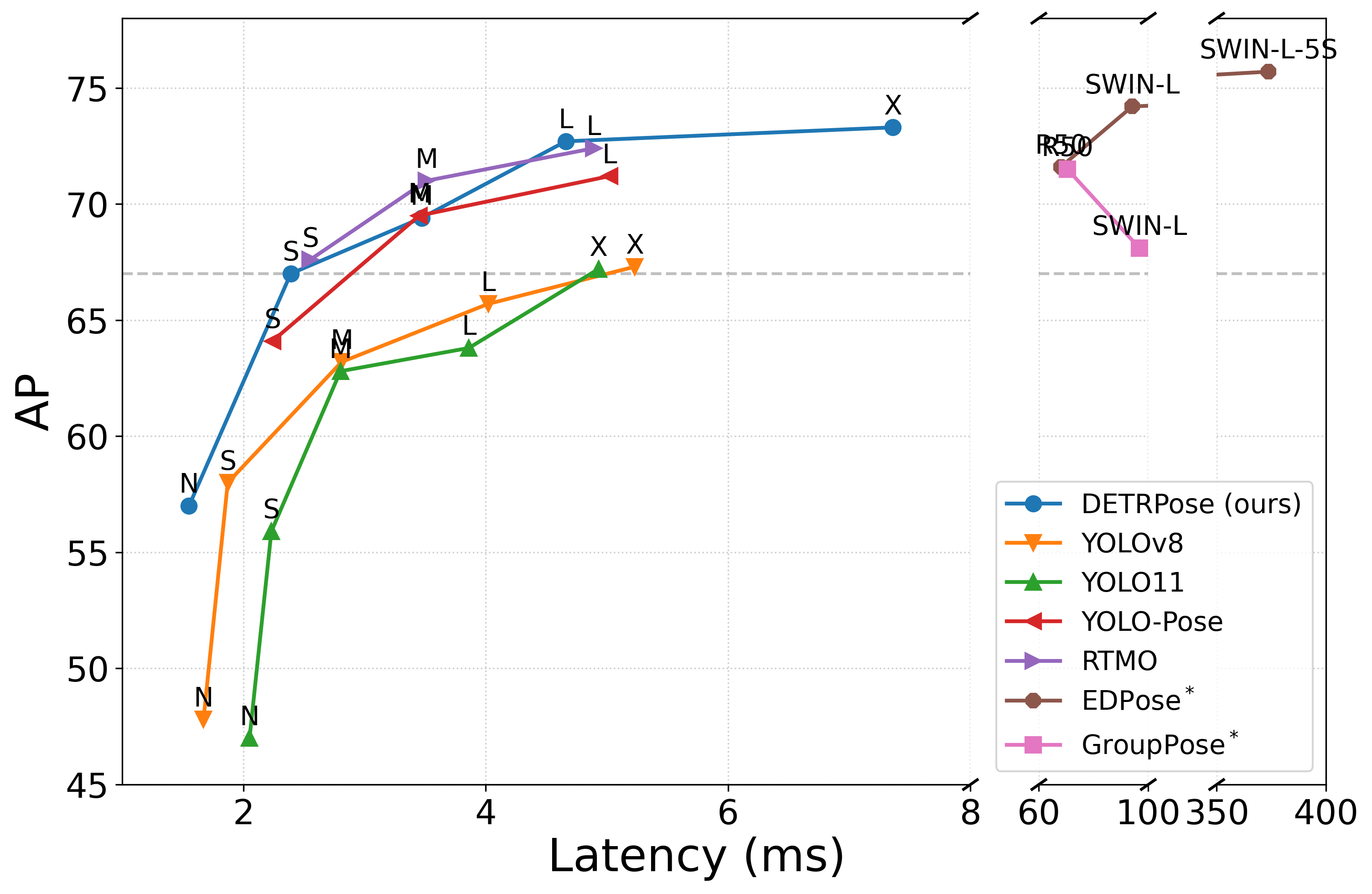} &
		\includegraphics[width=0.5\textwidth]{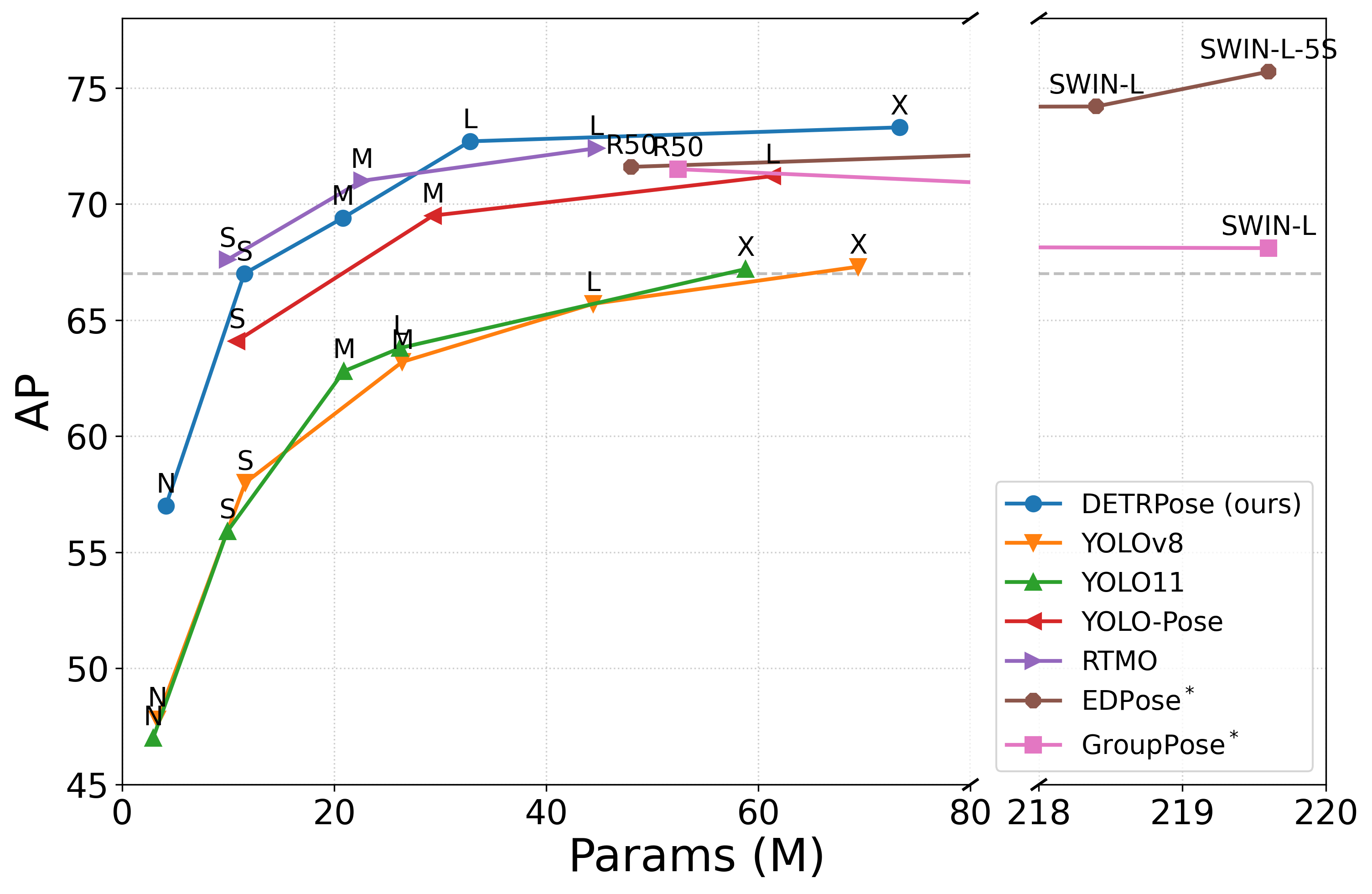} \\
		\multicolumn{2}{c}{a) COCO \texttt{val} dataset}\\
		\includegraphics[width=0.5\textwidth]{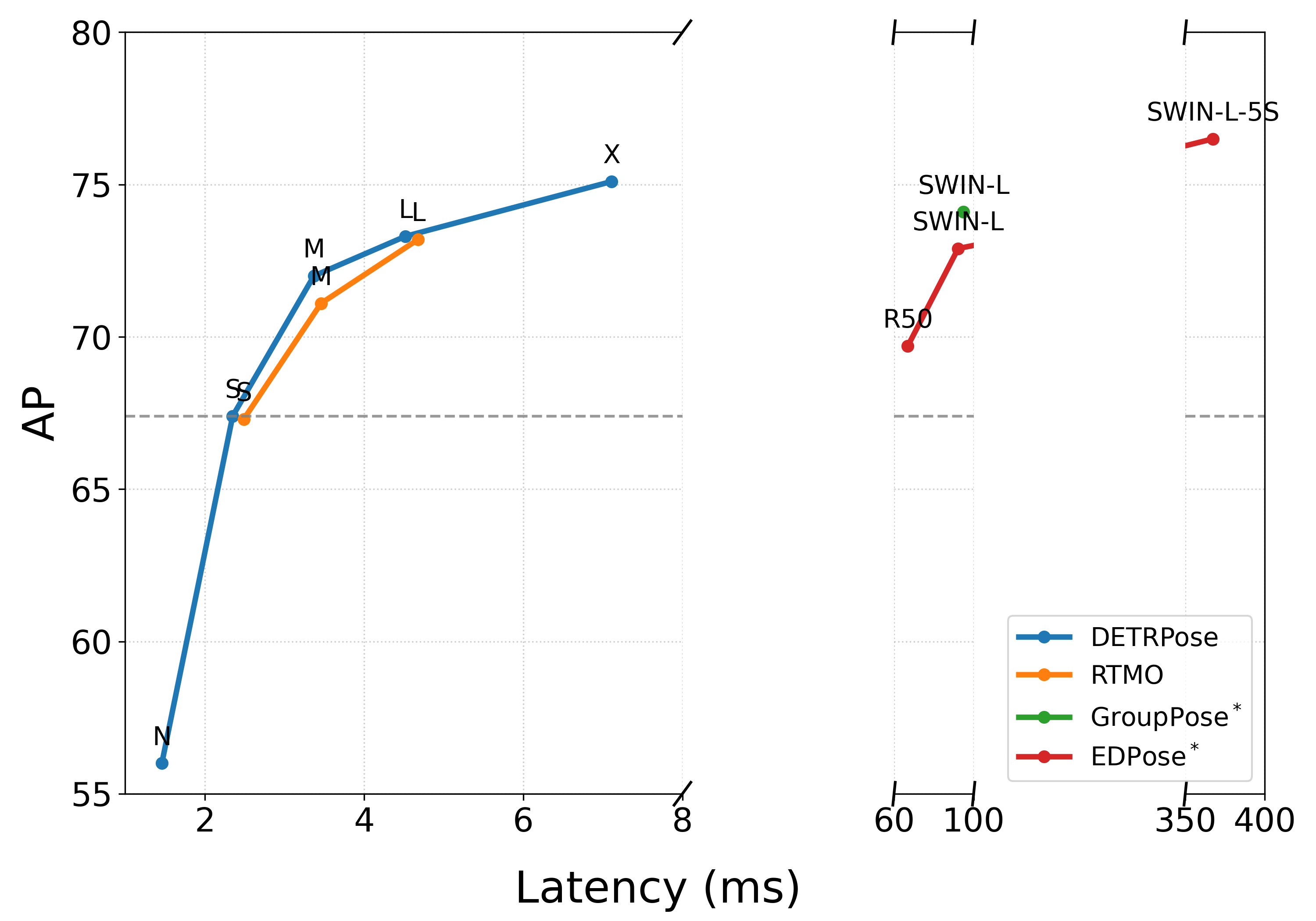} &
		\includegraphics[width=0.5\textwidth]{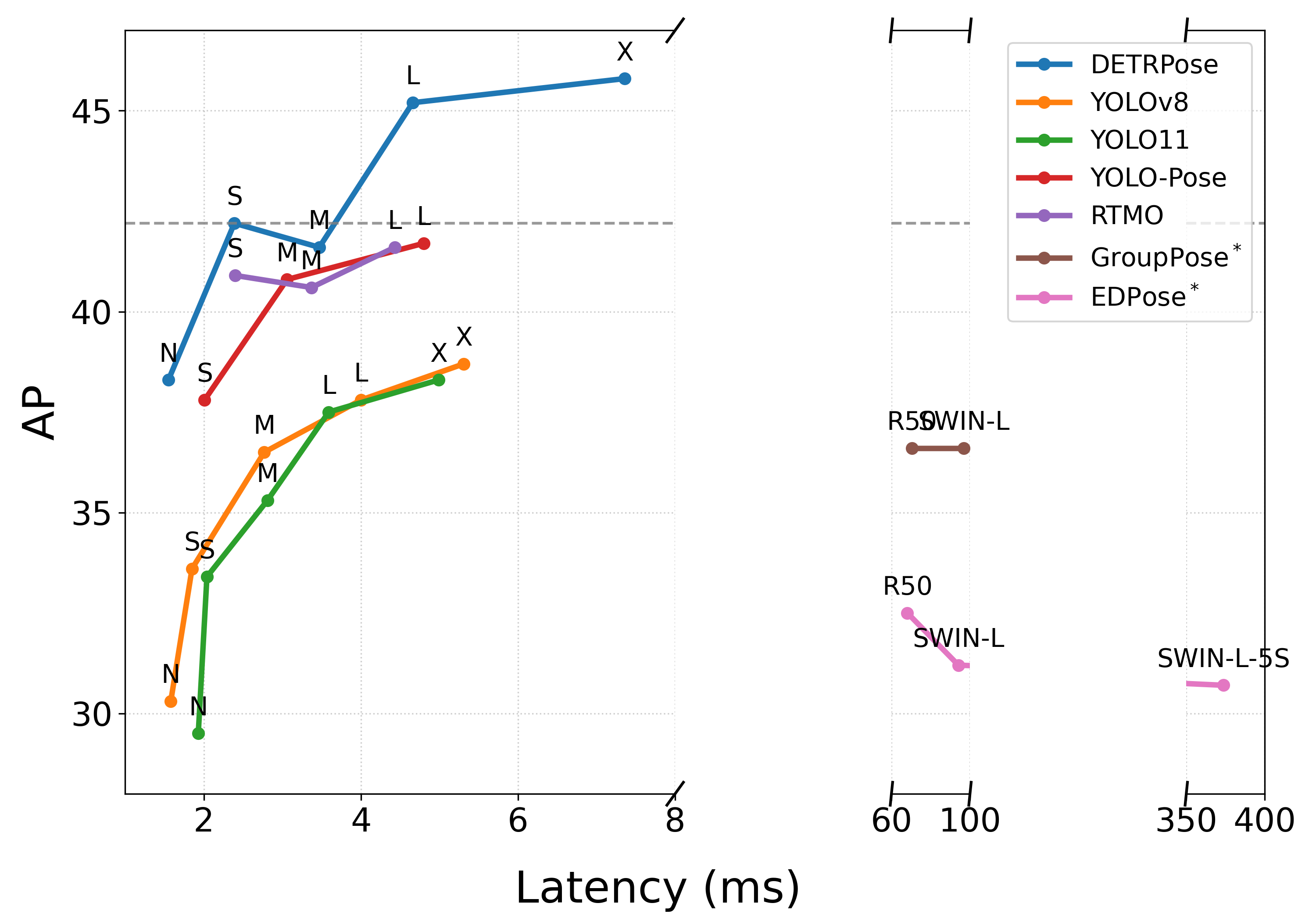} \\
		b) CrowdPose \texttt{test} dataset & c) OCHuman \texttt{test} dataset
		
	\end{tabular}
	
	\caption{Comparisons with other pose estimation models in mutiple datasets. We measure end-to-end latency using TensorRT FP16 on an NVIDIA RTX A5500 GPU.}
	\label{fig: metrics_comp}
\end{figure*}

The majority of fast MPPE methods can be categorized as either top-down or single-stage approaches. Top-down models first detect individuals, crop the object regions, and then process the cropped images to estimate keypoints. These models typically achieve higher accuracy than alternative methods, but their inference times increase linearly with the number of people in an image. In contrast, single-stage methods estimate human keypoints and subsequently group them for each person. Although bottom-up methods provide faster inference, they are generally less accurate because the models must simultaneously learn detection and pose estimation.

Recent developments in single-stage models (\eg YOLOv8/11-Pose \cite{yolov8, yolo11} and RTMO \cite{rtmo})  successfully adapted YOLO models for MPPE, providing a good trade-off between speed and precision. However, these methods rely on Non-Maximum Suppression (NMS) during inference, resulting in inconsistent latency.

To overcome the limitations associated with NMS, several studies have extended the DETR framework to pose estimation \cite{edpose, petr}. DETR-based models achieve consistent inference times and strong precision. However, prior transformer models have prioritized high precision over computational speed. Although real-time DETR object detectors have been developed \cite{rtdetr, dfine, deim}, no real-time adaptation of DETR for pose estimation exists, thereby restricting the broader application of the DETR framework.

This paper introduces DETRPose, the first real-time end-to-end transformer-based pose estimation model. To achieve low, consistent latency while maintaining high accuracy, several training innovations are presented, and the GroupPose decoder (which predicts keypoints from queries) is adapted to reduce processing time. A new denoising keypoint technique that uses Keypoint Similarity (KS) is proposed for constructing keypoint queries, significantly accelerating convergence. Additionally, the Keypoint Similarity VariFocal (KSVF) loss function and an auxiliary Pose-LQE (Localization Quality Estimation) classification layer are introduced to refine the quality of keypoint predictions and classification accuracy (same purpose as the LQE layers \cite{lqe} used in D-FINE\footnote{The use of LQE is not mentioned in the D-FINE paper, but it is found in D-FINE's decoder code from its official repository.} \cite{dfine}). Architecturally, the D-FINE strategy is adopted, and projection layers are removed to further reduce decoder latency.

The proposed approach achieves competitive results compared to previous methods while maintaining low and consistent latency. Specifically, DETRPose-L attains 71.2\% Average Precision (AP) on the COCO \texttt{test-dev2017} dataset with a latency of 4.66 ms, surpassing prior end-to-end methods. Compared to RTMO \cite{rtmo}, the current state-of-the-art model for real-time MPPE, DETRPose-L offers fewer parameters, reduced latency, faster convergence, a comparable AP score (only 0.4 points lower), and a higher AR score (+3.2 points). Against popular frameworks like YOLOv8-Pose and YOLO11-Pose, DETRPose-S matches the biggest version of these two YOLO models on the COCO \texttt{val} dataset (see \cref{fig: metrics_comp}a). On the CrowdPose \texttt{test} dataset, DETRPose-X exceeds expectations, surpassing all transformer models with the Swin-L backbone by at least 1 point in AP, while using significantly fewer parameters (73.3M vs. 210M). On the OCHuman dataset, DETRPose-S shows DETRPose's superior robustness on out-of-distribution datasets, surpassing RTMO-L (42.2 vs 41.6 in AP) while being $1.85\times$ faster (see \cref{fig: metrics_comp}c).

The primary contributions of this work are:
\begin{itemize}
	
	\item A new denoising technique for pose estimation. The proposed approach employs object Keypoint Similarity (KS) to construct both positive and negative keypoint queries.
	
	\item An auxiliary classification layer, the Pose-LQE layer, which enhances classification performance. Its lightweight design enables integration without increasing latency, addressing the challenge that DETRPose predicts keypoints rather than objects.
	
	\item A new loss function, the Keypoint Similarity VariFocal (KSVF) loss, that uses the Keypoint Similarity (KS) metric and the predicted score to promote the selection of high-quality object queries. 
	
	\item A modified GroupPose decoder to support real-time end-to-end MPPE for transformer-based models.
	
\end{itemize}

\section{Related Works}
Single-stage multi-person pose estimation methods can be categorized into two primary groups:
(i) non-end-to-end methods, and 
(ii) end-to-end methods.

\subsection{Non-End-to-End Methods}
Non-end-to-end methods \cite{directpose, fcpose, cid, inspose, yolopose, yolo11} can be further divided into single-stage and two-stage approaches. 
Single-stage methods \cite{yolo11, yolov8, rtmo} are fast because they estimate the keypoint instances in a single forward pass.
In contrast, two-stage approaches include both top-down and bottom-up methods. Top-down approaches generally achieve higher accuracy by first performing person detection and then pose estimation. During training, top-down methods detect humans by assigning bounding boxes to each person, then estimate pose within each bounding box. Conversely, bottom-up methods estimate all keypoints in the first stage and subsequently group keypoints into instances in the second stage. A key limitation of non-end-to-end methods is their reliance on hand-crafted processes such as Non-Maximum Suppression (NMS). Furthermore, two-stage processes introduce significant variability in inference time due to the need to process bounding boxes or group keypoints.

\subsection{End-to-End Methods}
Current end-to-end multi-person pose estimation (MPPE) models utilize the DETR \cite{detr} framework to eliminate the need for NMS. PETR \cite{petr} is the first DETR-based model for pose estimation and employs two decoders: a pose decoder and a joint decoder. The pose decoder coarsely detects human instances by estimating their poses, while the joint decoder promotes interaction between keypoints of the same person, resulting in more precise estimations. Subsequently, QueryPose \cite{querypose} and ED-Pose \cite{edpose} adopt a top-down approach by incorporating a human detector decoder. Notably, QueryPose implements two parallel branches: a bounding box detector and a pose detector. 
ED-Pose further advances these developments by employing sequential decoder layers, with initial layers dedicated to detection and subsequent layers estimating human keypoints. In contrast, GroupPose \cite{grouppose} utilizes group self-attention to eliminate the need for detection-only decoder layers. Group self-attention comprises two sequential self-attention modules: the first operates among keypoint queries within the same human instance, and the second operates among queries of the same keypoint type across different instances. Although these models demonstrate effectiveness, they do not fully exploit DETR's capabilities for pose estimation. Notably, no prior work has applied a denoising technique to keypoints, despite these being the primary instances being estimated.

\begin{figure}[!t]
	\centering
	\begin{tabular}{ccc}
		\includegraphics[width=0.2\columnwidth]{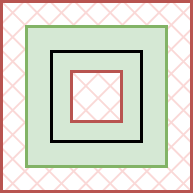}
		&&
		\includegraphics[width=0.2\columnwidth]{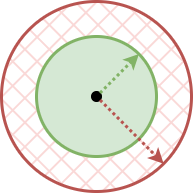} \\
		a) Denoising Box \cite{dndetr, dino} &\phantom{aaaaaaaaaa}& b) Denoising Keypoint (ours) 
	\end{tabular}
	\caption{Denoising techniques. \textbf{Left}.  Denoising box \cite{dndetr, dino} produced by shifting and scaling the ground-truth box,  used for object detection. \textbf{Right}. Denoising keypoint consists of shifting the ground-truth keypoint. The ground-truth instances are in black. The positive and negative queries are located within the green and red areas, respectively.}
	\label{fig:denoising}
\end{figure}

\subsection{Denoising Techniques}
DETR-based methods use query vectors to propose candidate instances for bounding box, keypoint, or mask estimation \cite{detr, edpose, maskdino}. Many works have focused on developing strong queries by associating them with specific spatial positions to improve results \cite{anchor, dabdetr}. However, none of them focus on the Hungarian matcher (the algorithm that enables DETR to perform end-to-end inference), which is unstable, especially during early training.

The denoising strategy was initially introduced in DN-DETR \cite{dndetr}, and later improved in DINO \cite{dino}, to stabilize the Hungarian Matcher. The denoising strategy generates both positive and negative queries. It trains the model to accept positive queries with bounding boxes close to the ground truth (based on their classification score) and to reject negative queries whose boxes are far from the ground truth. As shown in \cref{fig:denoising}a, DETR-based object detectors generate these two types of queries for object detection. For each ground-truth bounding box (black square), its center and dimensions are randomly shifted or scaled to build a positive (green area) or negative (red area) query, using the width and height of the ground-truth. 

In practice, ED-Pose \cite{edpose} applies the DINO denoising strategy to the detection layers. It then removes the denoising queries, keeping only the selected top-k queries for the human-to-keypoints layers. GroupPose \cite{grouppose} does not employ denoising, as it estimates only keypoint locations. Interestingly, previous pose estimation models have not applied denoising to keypoints, despite many DETR models relying on this approach for optimal performance \cite{rtdetr, dfine, deim}.

Unlike a bounding box that has 4 points (corners), keypoints are single points. This difference makes it challenging to shift or translate keypoints. Additionally, not all keypoints belong to the same joint type. To address these challenges, we propose a novel denoising keypoint strategy that uses the keypoint similarity to generate both positive and negative queries (see \ref{fig:denoising}b).

\begin{figure*}[!t]
	\centering
	\includegraphics[width=0.9\textwidth]{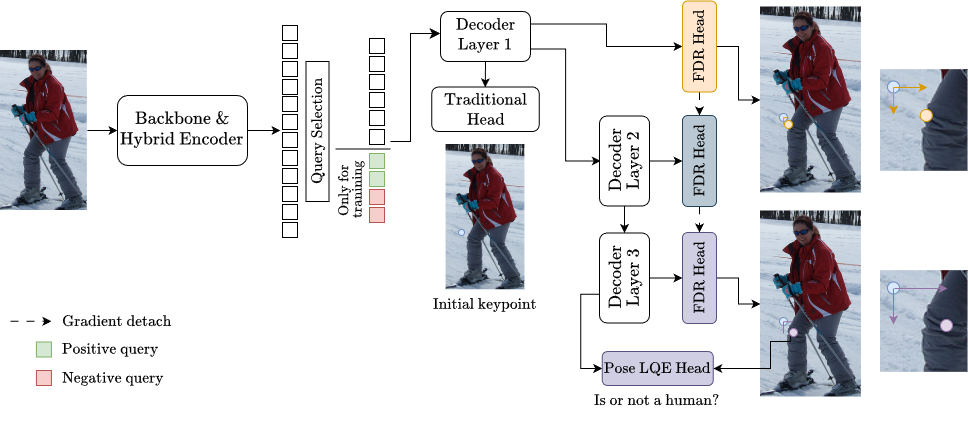}
	\caption{DETR-POSE architecture.
		All keypoints are estimated simultaneously.
		For simplicity, we present the computation for a single person. The arrows in the right-side images (knee) represent the vertical and horizontal offsets generated by the FDR heads.}
	\label{fig:model_overview}
\end{figure*}

\section{Methodology}
\Cref{fig:model_overview} shows the DETRPose architecture. It consists of a hierarchical backbone, a hybrid encoder, and a transformer decoder. DETRPose estimates all keypoints for every human instance simultaneously by employing group self-attention in each decoder layer. We use the backbone and encoder from D-FINE, initialize them with D-FINE's Object365 weights. We introduce a new training strategy named Denoising Keypoints strategy to accelerate convergence and to enhance overall model performance. We also develop the Pose-LQE head to refine classification scores and the Keypoint Similarity VariFocal loss to improve query selection quality. 

\subsection{Denoising Keypoints}
We next describe our denoising keypoint strategy. We begin by describing the keypoint similarity. Let $\kappa>0$ denote a fall-off parameter that will allow us to vary our similarity metric for each type of keypoint. The $\kappa$ values are taken from the  COCO  and CrowdPose repositories. Let $s$ be the object scale, and $0\leq s^2 \leq HW$, the human's segmented area. $H$ and $W$ are the image dimensions.
Then, using the distance $d$ between the ground-truth and a sampling keypoint, we define the keypoint similarity ($\mathrm{KS}$) as:

\begin{equation}
	\mathrm{KS} = \exp (-d^2/(2s^2\kappa^2)). \label{eq:ks}
\end{equation}

In order to derive bounds for $\mathrm{KS}$, we note that the sampling keypoint $\hat{p}$ can be written as a variation of the ground-truth keypoint $p$ given by:
\begin{equation}
	\hat{p} = p + \alpha_{pose}\overrightarrow{n}
	\label{eq:phat}
\end{equation}
where $\alpha_{\text{pose}}\geq 0$ is a non-fixed scaling factor and $\overrightarrow{n}$ is a random unit vector. In order to estimate $\alpha_{\text{pose}}$ for generating positive and negative queries, we note that $d(p, \hat{p}) = \alpha_{pose}$, and substitute it in \cref{eq:ks} to get:
\begin{equation}
	\alpha_{\text{pose}}^2 = -2\ln(\mathrm{KS})s^2\kappa^2.
	\label{eq:pose}
\end{equation} 
\begin{equation}
	\alpha_{\text{pose}} 
	= s\kappa\sqrt{-2\ln(\mathrm{KS})}.
	\label{eq:pose_final}
\end{equation}

Next, we describe how we generate positive and negative queries. We sample values for $\mathrm{KS}$ from a uniform distribution $\mathcal{U}$, then substitute them into \eqref{eq:pose_final} to get $\alpha_{\text{pose}}$. This $\alpha_{\text{pose}}$ value sets the magnitude of the shift. We generate the sampling keypoint, $\hat{p}$ , by translating the ground-truth keypoint, $p$, along a random unit vector $\overrightarrow{n}$ as in \eqref{eq:phat}. For positive queries, we sample $\mathrm{KS}$ from $\mathcal{U}(0.5, 1)$. For negative queries, we sample $\mathrm{KS}$ from $\mathcal{U}(0.1, 0.5)$.

During training, we apply an attention mask that \textit{i)} allows the denoising groups to interact with the prediction group (but not vice-versa), and \textit{ii)} that blocks cross interaction between denoising groups. We used a fixed number of 100 positive and 100 negative queries. For the CrowdPose dataset \cite{crowdpose}, we define the segmented area as $s^2 = \text{A}_\text{bbox} \cdot 0.53$ where  $\text{A}_\text{bbox}$ is the area of the bounding box, and the $0.53$ is taken from the CrowdPose's  evaluation code.

\subsection{Query Selection}
We use a linear layer processes the encoder’s feature maps to get the top-$N$ pixels with the highest probability of containing an instance query. Then, another linear layer generates $K$ keypoints for each selected pixel (giving a total of $K\times N$ keypoints).  Then, we redefine the instance query as the mean value of the $K$ keypoints for each human instance. Overall, the decoder receives $N\cdot (K+1)$ queries in total.

\subsection{Decoder}
Each decoder layer includes a within-instance self-attention layer, where keypoints from the same human instance interact. Then, an across-instance self-attention layer enables keypoints of the same joint to interact with one another. To separate negative, positive, and prediction instances, we apply a mask attention in the across-instances self-attention layer. Finally, a deformable attention mechanism allows all queries to interact with the encoder's feature maps.

Our decoder introduces two critical architectural enhancements over GroupPose: the integration of D-FINE's pre-pose and Fine-Grained Distribution Refinement (FDR) layers, and our novel Pose Localization Quality Estimation (Pose-LQE) layer, which enhances classification scores.

\subsubsection{Pose-LQE Layer}
\begin{figure}[!t]
	\includegraphics[width=\columnwidth]{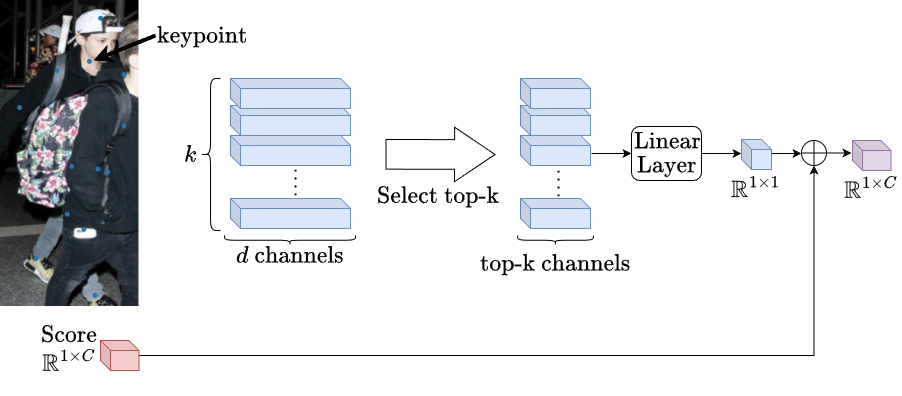}
	\caption{Pose-LQE layer for improving classification accuracy.}
	\label{fig:pose_lqe}
\end{figure}

The Pose-LQE layer extracts content information from the encoder’s highest-spatial feature map at each keypoint. It produces a vector for each keypoint and keeps only the top-k channels. A subsequent linear layer processes these values and combines them with those from traditional classification layers to produce a classification score. This approach creates a more confident classification score by combining information from the output queries, the predicted keypoint locations, and content features. \cref{fig:pose_lqe} illustrates this procedure.


\subsection{Keypoint Similarity VariFocal Loss}
Since DETRPose does not estimate bounding boxes, it can not be trained using an IoU-based loss. We propose the Keypoint Similarity VariFocal ($\mathrm{KSVF}$) loss defined as:
\begin{equation}
	\mathrm{KSVF}(q, \tilde{c}) =  \begin{cases}
		-q(q \log (\tilde{c}) + (1-q)\log(1-\tilde{c})), & q > 0\\
		-\alpha \tilde{c}^\gamma\log(1-\tilde{c}),   & q=0
	\end{cases}
\end{equation}
where $q$ is the object keypoint similarity (OKS), $\tilde{c}$ denotes the predicted classification score, and $\alpha$ and $\gamma$ are constants that control
the loss function for low-similarity cases.
We set $\alpha=0.25$ and $\gamma = 2.0$ during training.

Our $\mathrm{KSVF}$ is an extension of the VariFocal loss \cite{varifocalnet}, which is used in RT-DETR and D-FINE. The main purpose of $\mathrm{KSVF}$ is to reduce false-positive predictions by making the network focus on the predictions with high object keypoint similarity. A second benefit is that we can reduce the number of predicted persons because $\mathrm{KSVF}$ produces higher quality predictions. Since pose estimation transformers treat each keypoint as a query, reducing the number of predicted humans significantly reduces the decoder's FLOPs.
\begin{table}[!t]
	\centering
	\caption{\label{tab:parameters}Parameter for DETRPose family.}
	\begin{tabular}{l| c c c c}
		\toprule
		&DETRPose-S & \phantom{a}DETRPose-M\phantom{a}  & DETRPose-L & \phantom{a}DETRPose-X\phantom{a}\\
		\midrule
		\# decoder layers & 3 & 4 & 6 & 6\\
		\# human instances \phantom{a} & 60 & 60 & 60 & 60 \\
		hidden dim. & 256 & 256 & 256 & 384 \\
		backbone & HGNetv2-B0 & HGNetv2-B2 & HGNetv2-B4 & HGNetv2-B5\\
		epochs & 96 & 60 & 48 & 48\\
		lr & 1e-4 & 1e-4 & 1e-4 & 1e-4\\
		backbone lr & 1e-4 & 1e-5 & 1e-5 & 5e-5\\
		barch size & 16 & 16 & 16 & 16\\
		\bottomrule 
	\end{tabular}
\end{table}

\section{Results}

\subsection{Datasets}
We conduct experiments on two standard benchmarks: COCO \cite{coco}, CrowdPose \cite{crowdpose}, and OCHuman \cite{ochuman}. The COCO dataset includes over 150K images and 250K person instances annotated with 17 keypoints. CrowdPose, a more challenging dataset due to its highly crowded and occluded scenes, contains approximately 20K images and 80K person instances. The OCHuman dataset is only for testing and includes 5081 images and 10375 person instances with 17 keypoints.

We train DETRPose on the COCO \texttt{train} set and report results on the COCO \texttt{val} and \texttt{test-dev} sets, as well as the OCHuman \texttt{test} set.
For CrowdPose, we use the \texttt{train-val} set for training and evaluate on the \texttt{test} set.

\subsection{Model Variants}
We develop four DETRPose models of varying sizes: DETRPose-S, DETRPose-M, DETRPose-L, and DETRPose-X. \cref{tab:parameters} provides information about the decoder parameters, backbone name, and training hyperparameters. Each DETRPose variant adopts the same backbone and encoder architecture as its corresponding D-FINE counterpart \cite{dfine}.

\subsection{Implementation Details}
\label{sec:implementation}
All DETRPose variants are trained using the AdamW optimizer with a batch size of 16 on 4 Tesla V100 GPUs (cloud service). For data augmentation, we adopt the DEIM \cite{deim} protocol, which includes horizontal flipping, color jittering, mosaic composition, mix-up, and multi-scale sampling, but we do not use its cosine annealing strategy. Following the transfer learning strategy of RTMO \cite{rtmo} (see the supplementary material for more information), we initialize our backbone and encoder with D-FINE Object365 pretrained weights \cite{dfine}.

We use average precision (AP) to measure and compare model performance. During validation, input images are resized to $640 \times 640$. For ED-Pose and GroupPose, we apply letterboxing with a fixed input size of $800\times1300$ to preserve the image aspect ratio and provide a padding mask as an additional input. We report latency using TensorRT with FP16 precision on a local Nvidia RTX A5500. For a fair and consistent comparison, we use the benchmarking tool developed by Roboflow \cite{rfdetr}, which reports latency as the median rather than the mean.  Since this work was done before YOLO26's release \cite{yolo26}, we do not compare to it in the main paper but do so in the supplemental material.  

\subsection{Comparisons}

\begin{table}[!t]
	\centering
	\caption{\label{tab:coco-test}Performance comparison on the COCO \texttt{test-dev} dataset. The model with $^\dagger$ is trained without transfer learning. Times with *  resize are computed using an input size of $800\times1300$.}
	\resizebox{\columnwidth}{!}{
		\begin{tabular}{l c c c c ccc cc c}
			\toprule
			Method & Backbone & Epochs & \phantom{a}\#Params\phantom{a} & Time (ms) &\phantom{a}AP\phantom{a} & $\textrm{AP}_{50}$ & \phantom{a}$\textrm{AP}_{75}$\phantom{a} & $\textrm{AP}_{M}$ & \phantom{a}$\textrm{AP}_{L}$\phantom{a} & AR\phantom{a}\\
			\midrule
			\multicolumn{11}{c}{ Non-end-to-end methods}\\
			\midrule
			KAPAO-S~\cite{kapao} & CSPNet & 500 & 12.6M & - & 63.8 & 88.4 & 70.4 & 58.6 & 71.7 & 71.2\\
			KAPAO-M~\cite{kapao}  & CSPNet & 500 & 35.8M & - & 68.8 & 90.5 & 76.5 & 64.3 & 76.0 & 76.3\\
			KAPAO-L~\cite{kapao}  & CSPNet & 500 & 77.0M & - & 70.3 & \textit{91.2} & 77.8 & 66.3 & 76.8 & 77.7\\
			YOLO-Pose-S~\cite{yolopose} & CSPDarknet  & 300 & 10.8M & 2.24 & 63.2 & 87.8 & 69.5 & 57.6 & 72.6 & 67.6\\
			YOLO-Pose-M~\cite{yolopose}  & CSPDarknet & 300 & 29.3M & 3.45 & 68.6 & 90.7 & 75.8 & 63.4 & 77.1 & 72.8\\
			YOLO-Pose-L~\cite{yolopose}  & CSPDarknet & 300 & 61.3M & 5.02 & 70.2 & 91.1 & 77.8 & 65.3 & 78.2 & 74.3\\
			RTMO-S \cite{rtmo} & CSPDarknet & 600 & 9.9M & 2.55 & 66.9 & 88.8 & 73.6 & 61.1 & 75.7 & 70.9\\
			RTMO-M \cite{rtmo} & CSPDarknet & 600 & 22.6M & 3.51 & 70.1 & 90.6 & 77.1 & 65.1 & 78.1 & 74.2\\
			RTMO-L \cite{rtmo} & CSPDarknet & 600 & 44.8M & 4.89 & \textit{71.6} & 91.1 & \textit{79.0} & \textit{66.8} & 79.1 & 75.6\\
			\midrule
			\multicolumn{11}{c}{End-to-end methods with Swin-L backbone}\\
			\midrule
			PETR~\cite{petr} & Swin-L & 100 & 213.8M & - & 70.5 & 91.5 & 78.7 & 65.2 & 78.0 & -\\
			QueryPose\cite{querypose} & Swin-L & 79 & - & - & 73.3 & 91.3 & 79.5 & 68.5 & 81.2 & -\\
			ED-Pose~\cite{edpose} & Swin-L & 60 & 218.4M & 94.11* & 72.7 & 92.3 & 80.9 & 67.6 & 80.0 & -\\
			GroupPose* \cite{grouppose} & Swin-L & 60 & 219.6M & 96.77* & 74.8 & 91.6 & 82.1 & 69.4 & 83.0 & -\\
			\midrule
			\multicolumn{11}{c}{End-to-end methods}\\
			\midrule
			PETR*~\cite{petr} & ResNet-50 & 100 & 43.7M & - & 67.6 & 89.8 & 75.3 & 61.6 & 76.0 & -\\
			QueryPose*\cite{querypose} & ResNet-50 & 40 & - & - & 68.7 & 88.6 & 74.4 & 63.8 & 76.5 & -\\
			ED-Pose*~\cite{edpose} & ResNet-50 & 60 & 48.0M & 68.10* & 69.8 & 90.2 & 77.2 & 64.3 & 77.4 & -\\
			GroupPose* \cite{grouppose} & ResNet-50 & 60 & 52.4M & 70.44* & 70.2 & 90.5 & 77.8 & 64.7 & 78.0 & - \\
			DETRPose-L$^\dagger$ (ours) & ResNet-50 & 48 & 42.7M & 4.98& 68.7 & 89.3 & 75.2 & 62.6 & 78.3 & 75.8 \\
			DETRPose-S (ours)& HGNetv2-B0 & 96 & 11.5M & 2.39 & 66.5 & 87.4 & 72.5 & 59.8 & 76.8 & 73.0\\
			DETRPose-M (ours)& HGNetv2-B2 & 60 & 20.8M & 3.47 & 68.5 & 88.6 & 74.6 & 62.1 & 78.3 & 74.6\\
			DETRPose-L (ours)& HGNetv2-B4 & 48 & 32.8M & 4.66 & 71.2 & \textit{91.2} & 78.1 & 65.8 & \textit{79.9} & \textit{78.1} \\
			DETRPose-X (ours) & HGNetv2-B5 & 48 & 73.3M & 7.36 & \textbf{72.2} & \textbf{91.4} & \textbf{79.3} & \textbf{67.0} & \textbf{80.6} & \textbf{78.8} \\
			\bottomrule
		\end{tabular}
	}
\end{table}

\begin{table}[!t]
	\centering\caption{\label{tab:coco-VAL}Performance comparison  on the COCO \texttt{val} dataset. Best and second results are in \textbf{bold} and \textit{italic}, respectively.}
	\resizebox{\columnwidth}{!}{
		\begin{tabular}{l c c ccc ccc c}
			\toprule
			Method & \# Params & \phantom{a}Time\phantom{a} & FLOPs & \phantom{a}AP\phantom{a} & $\textrm{AP}_{50}$ & \phantom{a}$\textrm{AP}_{75}$\phantom{a} & $\textrm{AP}_{M}$ & \phantom{a}$\textrm{AP}_{L}$\phantom{a} & AR\phantom{a}\\
			& & (ms) & (G) & & & & & & \\
			\midrule
			YOLO-Pose-S~\cite{yolopose} & 10.8M & 2.24 & 18 & 64.1 & 87.1 & 70.2 & 57.9 & 74.1 & 68.2 \\
			YOLO-Pose-M~\cite{yolopose}  & 29.3M & 3.45 & 48 & 69.5 & 89.9 & 76.2 & 63.7 & 78.9 & 73.3\\
			YOLO-Pose-L~\cite{yolopose} &61.3M & 5.02 & 98 & 71.2 & 90.1 & 78.2 & 65.6 & 80.1 & 74.9\\
			RTMO-S \cite{yolo11} & 9.9M & 2.55 & 12 & 67.6 & 87.8 & 73.6 & 61.4 & 77.3 & 71.5\\
			RTMO-M \cite{yolo11} & 22.6M & 3.51 & 32 & 71.0 & 89.5 & 77.8 & 65.3 & 79.8 & 74.8\\
			RTMO-L \cite{yolo11} & 44.7M & 4.89 & 68 & 72.4 & 89.9 & 78.8 & 67.1 & 81.0 & 76.2 \\
			YOLOv8-S \cite{yolov8} & 11.6M & 1.87 & 30 & 58.0 & 82.4 & 64.7 & 49.8 & 70.0 & 66.7\\
			YOLOv8-M \cite{yolov8} & 26.4M & 2.81 & 81 & 63.2 & 85.6 & 70.8 & 56.6 & 73.1 & 71.8\\
			YOLOv8-L \cite{yolov8} & 44.4M & 4.02 & 169 & 65.7 & 86.6 & 73.0 & 59.4 & 75.1 & 74.2\\
			YOLOv8-X \cite{yolov8} & 69.4M & 5.23 & 263 & 67.3 & 87.5 & 75.0 & 61.5 & 76.1 & 75.7\\
			YOLO11-S \cite{yolo11} & 9.9M & 2.23 & 23 & 55.9 & 81.2 & 62.1 & 47.8 & 67.6 & 66.1\\
			YOLO11-M \cite{yolo11} & 20.9M & 2.80 & 72 & 62.8 & 85.4 & 70.6 & 55.9 & 71.4 & 72.2\\
			YOLO11-L \cite{yolo11} & 26.2M & 3.86 & 91 & 63.8 & 86.2 & 71.4 & 58.2 & 72.5 & 73.3\\
			YOLO11-X \cite{yolo11} & 58.8M & 4.93 & 203 & 67.2 & 87.5 & 74.8 & 62.3 & 75.5 & 76.3\\
			\midrule
			\multicolumn{10}{c}{End-to-end methods with Swin-L backbone}\\
			\midrule
			ED-Pose-SWIN-L*~\cite{edpose} & 218.4M & 94.11 & 1954 & 74.2 & 91.8 & 81.4 & 68.3 & 82.8 & 83.0 \\
			ED-Pose-SWIN-L-5S*~\cite{edpose} & 218.7M & 373.72 & 3061& 75.7 & 92.7 & 83.1 & 70.0 & 83.8 & 84.2 \\
			GroupPose-SWIN-L* \cite{grouppose} & 219.6M & 96.77 & 1977 & 68.1 & 89.8 & 74.7 & 60.3 & 79.2 & 78.2 \\
			\midrule
			\multicolumn{10}{c}{End-to-end methods}\\
			\midrule
			ED-Pose-R50*~\cite{edpose} & 48.0M & 68.10 & 591 & 71.6 & 90.1 & 78.0 & 66.0 & 80.0 & 81.2 \\
			GroupPose-R50* \cite{grouppose} & 52.4M & 70.44 & 614 & 71.5 & 89.5 & 78.8 & 66.2 & 79.8 & 81.5 \\
			DETRPose-S (ours) & 11.5M & 2.39 & 33 & 67.0 & 88.1 & 72.9 & 60.4 & 77.3 & 73.5\\
			DETRPose-M (ours) & 20.8M & 3.47 & 67 & 69.4 & 89.8 & 75.5 & 63.1 & 79.1 & 75.5\\
			DETRPose-L (ours) & 32.8M & 4.66 & 107 & \textit{72.7} & \textbf{91.0} & \textit{79.2} & \textit{66.7} & \textit{82.2} & \textit{78.7}\\
			DETRPose-X (ours) & 73.3M & 7.36 & 240 & \textbf{73.3} & \textit{90.5} & \textbf{79.4} & \textbf{67.5} & \textbf{82.7} & \textbf{79.4}\\
			\bottomrule
		\end{tabular}
	}
\end{table}

\begin{table*}[!t]
	\centering\caption{\label{tab:crowdpose_test}Performance comparison on the CrowdPose \texttt{test} dataset. Models with * use input size of $800\times1300$. Best and second results are in \textbf{bold} and \textit{italic}.}
	\resizebox{\textwidth}{!}{
		\begin{tabular}{l c c c c ccc cc c}
			\toprule
			Method  & Epochs & \phantom{a}\# Params\phantom{a} &  Time & FLOPs & \phantom{a}AP\phantom{a} & $\textrm{AP}_{50}$ & \phantom{a}$\textrm{AP}_{75}$\phantom{a} & $\textrm{AP}_{E}$ & \phantom{a}$\textrm{AP}_{M}$\phantom{a} & $\textrm{AP}_{H}$\phantom{a}\\
			& & & (ms) & (G) & & & & & & \\
			\midrule
			\multicolumn{11}{c}{ Non-end-to-end methods}\\
			\midrule
			RTMO-S \cite{rtmo}  & 700 & 9.8 M & 2.49  & 12 & 67.3 & 88.2 & 72.9 & 73.7 & 68.2 & 59.1\\
			RTMO-M \cite{rtmo}  & 700 & 22.4M & 3.46 & 32 & 71.1 & 87.7 & 77.1 & 77.4 & 71.9 & 63.4 \\
			RTMO-L \cite{rtmo}  & 700 & 44.5M & 4.68 & 68 & 73.2 & 90.7 & 79.3 & 79.2 & \textit{74.1} & 65.3\\
			\midrule
			\multicolumn{11}{c}{End-to-end methods with Swin-L backbone}\\
			ED-Pose-SWIN-L*~\cite{edpose}  & 80 & 218.3M & 92.22 & 1940 & 72.9 & 91.1 & 79.8 & 80.5 & 73.8 & 63.8\\
			ED-Pose-SWIN-L-5S*~\cite{edpose}  & 80 & 218.6M & 367.45 & 3048 & 76.5 & 92.6 & 83.3 & 83.0 & 77.3 & 68.3\\
			GroupPose -SWIN-L*\cite{grouppose}  & 80 & 219.7M & 94.89 & 1969 & 74.1 & 91.3 & 80.4 & 80.8 & 74.7 & 66.4\\
			\midrule
			\multicolumn{11}{c}{End-to-end methods}\\
			\midrule
			ED-Pose-R50*~\cite{edpose}  & 80 & 48.0M & 66.91 & 578 & 69.7 & 89.0 & 75.8 & 77.7 & 70.6 & 60.9 \\
			DETRPose-S (ours) & 156 & 11.5M & 2.35 & 31 & 67.4 & 88.6 & 72.9 & 74.7 & 68.1 & 59.3\\
			DETRPose-M (ours) & 72 & 20.7M & 3.37 & 65 &72.0 & 91.0 & 77.8 & 78.6 & 72.6 & 64.5\\
			DETRPose-L (ours) & 48 & 32.7M & 4.52 & 104 & \textit{73.3} & \textit{91.6} & \textit{79.4} & \textit{79.5} & 74.0 & \textit{66.1} \\
			DETRPose-X (ours) & 48 & 73.3M & 7.11 & 232 & \textbf{75.1} & \textbf{92.1} & \textbf{81.3} & \textbf{81.3} & \textbf{75.7} & \textbf{68.1}\\
			\bottomrule
		\end{tabular}
	}
\end{table*}

\begin{table*}[!t]
	\centering\caption{\label{tab:oc_human}Performance comparison on the OCHuman dataset.  Models with * use input size of $800\times1300$.  Best and second results are in \textbf{bold} and \textit{italic}.}
		\begin{tabular}{l cc ccc c}
			\toprule
			Method & \# Params & \phantom{a}Time\phantom{a}  & \phantom{a}AP\phantom{a} & $\textrm{AP}_{50}$ & \phantom{a}$\textrm{AP}_{75}$\phantom{a} & $\textrm{AP}_{L} $\\
			& & (ms)  & & & & \\
			\midrule
			YOLO-Pose-S~\cite{yolopose} & 10.8M &  2.01 & 37.8 & 52.6 & 42.9 & 73.4\\
			YOLO-Pose-M~\cite{yolopose}  & 29.3M & 3.06 & 40.8 & 53.2 & 46.1 & 78.2\\
			YOLO-Pose-L~\cite{yolopose} & 61.3M & 4.80 & 41.7 & 53.2 & 47.0 & 79.4\\
			RTMO-S \cite{yolo11} & 9.9M & 2.40 & 40.9 & 53.5 & 45.8 & 77.3\\
			RTMO-M \cite{yolo11} & 22.6M & 3.37 & 40.6 & 51.9 & 45.4 & 79.7\\
			RTMO-L \cite{yolo11} & 44.7M &  4.43 & 41.6 & 51.9 & 46.1 & 81.0\\
			YOLOv8-S \cite{yolov8} & 11.6M & 1.85 & 33.6 & 49.3 & 37.4 & 68.3\\
			YOLOv8-M \cite{yolov8} & 26.4M & 2.77 & 36.5 & 50.3 & 40.8 & 72.5\\
			YOLOv8-L \cite{yolov8} & 44.4M & 4.00 & 37.8 & 51.1 & 42.4 & 74.4\\
			YOLOv8-X \cite{yolov8} & 69.4M & 5.31 & 38.7 & 51.1 & 43.3 & 76.3\\
			YOLO11-S \cite{yolo11} & 9.9M & 2.04 & 33.4 & 49.6 & 37.6 & 67.4\\
			YOLO11-M \cite{yolo11} & 20.9M & 2.81 & 35.3 & 49.7 & 39.9 & 71.5\\
			YOLO11-L \cite{yolo11} & 26.2M & 3.59 & 37.5 & 51.4 & 42.1 & 73.0\\
			YOLO11-X \cite{yolo11} & 58.8M & 4.99 & 38.3 & 51.1 & 43.2 & 75.0\\
			\midrule
			\multicolumn{7}{c}{End-to-end methods with Swin-L backbone}\\
			\midrule
			ED-Pose-SWIN-L*~\cite{edpose} & 218.4M & 94.11 & 31.2 & 38.8 & 34.7 & 89.4\\
			ED-Pose-SWIN-L-5S*~\cite{edpose} & 218.4M & 373.72 & 30.7 & 37.9 & 34.0 & 90.4\\
			GroupPose-SWIN-L* \cite{grouppose} & 219.6M & 96.77 & 36.6 & 45.7 & 41.1 & 88.6\\
			\midrule
			\multicolumn{7}{c}{End-to-end methods}\\
			\midrule
			ED-Pose-R50*~\cite{edpose} & 48.0M & 68.10 & 32.5 & 40.8 & 36.2 & 87.8\\
			GroupPose-R50* \cite{grouppose} & 52.4M & 70.44 & 36.6 & 45.2 & 40.5 & 88.7\\
			DETRPose-S (ours) & 11.5M & 2.39 & 42.2 & 54.3 & 47.5 & 85.2\\
			DETRPose-M (ours) & 20.8M & 3.47 & 41.6 & 53.0 & 46.9 & 86.2\\
			DETRPose-L (ours) & 32.8M & 4.66 & \textit{45.2} & \textit{55.4} & \textit{50.4} & \textbf{89.0}\\
			DETRPose-X (ours) & 73.3M & 7.36 & \textbf{45.8} & \textbf{55.7} & \textbf{51.1} & \textit{87.8}\\
			\bottomrule
		\end{tabular}
\end{table*}

\subsubsection{COCO Dataset Comparisons}
We provide comprehensive comparisons against several other methods using the COCO \texttt{test-dev} dataset in \cref{tab:coco-test}. Among the end-to-end models not using the Swin-L backbone,
DETRPose-L yields the best results while being $14.6\times$ faster. Compared with the second-best model, GroupPose \cite{grouppose}, DETRPose-L has 37\% fewer parameters (32.8M vs 52.4M).
Compared to the best non-end-to-end models, RTMO-L \cite{rtmo}, DETRPose-L has 27\% fewer parameters (32.8M vs 44.8M), 5\% lower latency (4.89 vs 4.66), 0.5\% lower AP score (71.2 vs 71.6), and 4\% higher AR score.
DETRPose-X , our biggest model, achieves competitive results against
end-to-end models with Swin-L backbones ($>$ 200M parameters). It is important to note that PETR, QueryPose, ED-Pose, and GroupPose resize the image so that its shortest side is 800 and preserve the original image aspect ratio, which explains why these models achieve better results than DETRPose-L with ResNet50.

We report a comparison on the COCO \texttt{val} dataset on \cref{tab:coco-VAL}. DETRPose-X surpasses all previous methods by at least 0.9 and 3.1 points in AP and AR, respectively. \cref{fig: metrics_comp}a shows that DETRPose-S has a similar AP to YOLOv8/11-L models while having 32\% lower latency and 40\%  fewer parameters. DETRPose-M matches YOLOv8/11-X-Pose in the AP score (69.4 vs 69.2 and 69.5), and has 64\% fewer parameters, and 21\% lower latency. GroupPose-Swin-L-5S's performance lags behind its official PyTorch results because this GroupPose variant was trained with 1 image per GPU across 16 GPUs.

\subsubsection{CrowdPose Dataset Comparisons}
We provide comprehensive comparisons on the CrowdPose dataset in \cref{tab:crowdpose_test} and \cref{fig: metrics_comp}b.
The results on CrowdPose are more significant, as it is a more challenging dataset. DETRPose-M surpasses the ED-Pose-R50 while having  50\% fewer parameters, and $16\times$ lower latency. Compared with non-end-to-end models, DETRPose models outperform their RTMO counterparts (S, M, and L) and converge faster. For example, DETRPose-S and -M require 156 and 72 epochs, which is $4.5\times$ and $9.7\times$ more than for RTMO counterparts.

DETRPose-X sets a new state-of-the-art record for real-time MPPE models, achieving a score of 75.1 on the AP metric.  DETRPose-X surpasses most transformer models equipped with Swin-L while having $3\times$ less parameters. Only ED-Pose-SWIN-L-5S beats us for 1.5 points; however, its full model was pretrained on the Object365 dataset \cite{objects365}, and uses 5 feature maps for prediction.
Clearly, DETRPose-X provides strong evidence that our proposed transformer-based method holds great promise for pose estimation.

\subsubsection{OCHuman Dataset Comparisons}
We provide comparisons on the OCHuman dataset in \cref{tab:oc_human} to evaluate models' robustness. For fair comparisons, all the models are trained on the COCO \texttt{train} set.  DETRPose-S surpasses all methods by at least 0.6 while being the 4th fastest model in \cref{tab:oc_human}.  The Big models, ED-Pose and GroupPose, yield the lowest results, indicating they overfit the COCO dataset.  \cref{fig: metrics_comp}c validates this statement since ED-Pose-R50 performs better than it bigger variations. In terms of model scaling, the difference in AP between RTMO-S and RTMO-L is 0.7, while it is 3.0 for DETRPose, indicating that DETRPose scales better than other models on out-of-distribution datasets.

\subsection{Ablation study}

\begin{table}[!t]
	\centering
	\caption{Ablation study using DETRPose-L on the COCO\texttt{val} set.}
	\label{tab:ab_study}
	\resizebox{\columnwidth}{!}{
		\begin{tabular}{l | c c c c c | c c c c}
			\toprule
			Model number & \phantom{a}\# Detections & \phantom{a}KSVF & \phantom{a}Pose-LQE\phantom{a} & DK\phantom{a} & Pretraining\phantom{a} & \phantom{a}mAP & \phantom{a}\# Params & \phantom{a}FLOPs (G) \phantom{a} & Latency (ms)\\
			\midrule
			\multicolumn{9}{c}{Component Analysis}\\
			\midrule
			Model 1\phantom{a} & 100 & & & & None & \phantom{a}69.9 & 32.8M & 121.6 & 5.24 \\
			Model 2\phantom{a} & 60 & & & & None & \phantom{a}68.9 & 32.7M & 107.1  & 4.58\\
			Model 3\phantom{a} & 60 & \checkmark & & & None & \phantom{a}69.9 & 32.7M & 107.1 & 4.58\\
			Model 4\phantom{a} & 60 & & \checkmark & & None & \phantom{a}69.5 & 32.8M & 107.1 & 4.66\\
			Model 5\phantom{a} & 60 & & & \checkmark & None & \phantom{a}70.0 & 32.7M & 107.1 & 4.58\\
			Model 6\phantom{a} & 60 & & \checkmark & \checkmark & None & \phantom{a}69.6 & 32.8M & 107.1 & 4.66\\
			Model 7\phantom{a} & 60 & \checkmark & & \checkmark & None &  \phantom{a}70.9 & 32.7M & 107.1 & 4.58\\
			Model 8\phantom{a} & 60 & \checkmark & \checkmark & & None & \phantom{a}70.5 & 32.8M & 107.1 & 4.66\\
			Model 9\phantom{a} & 60 & \checkmark & \checkmark & \checkmark & None & \phantom{a}71.5 & 32.8M & 107.1 & 4.66\\
			\midrule
			\multicolumn{9}{c}{Pretraining Analysis}\\
			\midrule
			Model 9\phantom{a} & 60 & \checkmark & \checkmark & \checkmark & None & \phantom{a}71.5 & 32.8M & 107.1 & 4.66\\
			Model 10\phantom{a} & 60 & \checkmark & \checkmark & \checkmark & Coco & \phantom{a}71.1 & 32.8M & 107.1 & 4.66\\
			Model 11\phantom{a} & 60 & \checkmark & \checkmark & \checkmark & Obj2Coco & \phantom{a}72.4 & 32.8M & 107.1 & 4.66\\
			Model 12\phantom{a} & 60 & \checkmark & \checkmark & \checkmark & Obj365 & \phantom{a}72.7 & 32.8M & 107.1 & 4.66\\
			\bottomrule
		\end{tabular}
	}
\end{table}

\subsubsection{Component Analysis}
We provide a step-by-step analysis for each proposed component in DETRPose in \cref{tab:ab_study}. For each step, we fully train DETRPose-L on the COCO \texttt{train} set and report results on the COCO \texttt{val} set. To initiate our study, we create RT-GroupPose by adding D-FINE’s backbone, encoder, and FDR heads to the original GroupPose (first row in \cref{tab:ab_study}), achieving an AP of 69.9. Reducing the number of instance queries from 100 to 60 (Model 2) lowers AP and GFLOPs.
Adding the proposed components individually (Models 3–5) improves performance, with the denoising strategy giving the largest gain. 

We explore the synergy between different components (Models 6–8). The KSVF loss works well with either the denoising strategy or Pose-LQE, achieving 70.9 and 70.5, respectively. However, pairing the denoising strategy with Pose-LQE performs worse than using either component alone. Training DETRPose-L with all three proposed components achieves an AP of 71.5 (Model 9).

\subsubsection{Pretraining Weights.}
DETRPose-L achieves a score of 71.5 without pretraining initialization. Initializing the model with DFINE’s COCO-pretrained weights lowers performance, while Object365 weights yield the best result at 72.7. These experiments show that initializing DETRPose with DFINE weights is unnecessary to surpass YOLOv8 and YOLO11.

\subsubsection{Denoising Keypoint Analysis}
To understand how the per-keypoint fall-off parameter $\kappa$ and the segmented area $s^2$ affect the denoising keypoint strategy, we run multiple experiments varying the available information and report our findings in \cref{tab:ab_denoise}. We start by assigning a constant $\kappa = 1/ (\text{\# keypoints})$ for each type of keypoint, and define the object area $s^2 = \text{A}_\text{bbox} \cdot 0.53$. Using the denoising strategy without prior knowledge of the dataset performs better than training with KSFVF and Pose-LQE (Model 8). Providing the official COCO $\kappa$ values adds 0.1, even without masks to extract the area value. Keeping $\kappa$ constant but computing the area from masks adds another 0.1. Using both COCO $\kappa$ values and masks (Model 16) gives the best result, with a final score of 71.5. These experiments show that DETRPose can be fine-tuned on custom 2D MPPE datasets that lack mask information (see \cite{sam, sam2} for how to address this problem). 

\begin{table}[!t]
	\centering
	\caption{Ablation study of the denoising keypoints on the COCO-\texttt{val} dataset. $\kappa$ and $s^2$ represent the per-keypoint fall-off parameter and the segmented area, respectevily.}
	\label{tab:ab_denoise}
	\begin{tabular}{l |cc | c}
		\toprule
		ID & $\kappa$ & $s^2$ & \phantom{a}AP\phantom{a} \\
		\midrule
		ID13 & \phantom{a}constant\phantom{a} & box  & 71.1 \\
		ID14 &coco     & box  & 71.2\\
		ID15 & constant & \phantom{a}mask\phantom{a} &  71.3 \\
		ID16 \phantom{a} & coco     & mask & 71.5 \\
		\bottomrule
	\end{tabular}
\end{table}

\subsubsection{Flexibility of DETRPose}
We modify the number of layers and the number of queries in DETRPose-L and report results in \cref{tab:detrpose_comp}. We use the DETRPose-L trained on the COCO dataset, and the results in this section do not require any additional training. First, we remove deeper layers from the decoder. We observe that using 4 layers results in a 1.2\% drop in AP score, while latency reduces by 13.5\% compared to the original architecture (6 decoder layers). To our surprise, when we use 3 layers in the decoder, the AP significantly drops. 

We observe that reducing the number of queries has a small effect on performance. One reason is that the average and maximum numbers of people in an image are significantly lower than 40.  We also evaluate DETR-Pose-L with 25 queries and 4 and 5 decoder layers (last two rows of \cref{tab:detrpose_comp}), achieving better results than YOLOv8/11-X while being faster than YOLOv8/11-L.

\begin{table}[t]
	\centering
	\caption{Study of the flexibility of  DETRPose-L on the COCO-\texttt{val} dataset.}
	\label{tab:detrpose_comp}
	\begin{tabular}{cc | c c c c}
		\toprule
		\# Queries & \phantom{a}\# Layers\phantom{a} & \phantom{a}GFLOPs\phantom{a} & \phantom{a}\# Params\phantom{a} & \phantom{a}AP\phantom{a} & \phantom{a}Latency (ms)\phantom{a}\\
		\midrule
		\multicolumn{6}{c}{Varying the \# of decoder layers}\\
		\midrule
		60 &  6 & 107 & 32.8M & 72.7 & 4.66\\
		60 &  5 & 104 & 31.4M & 72.6 & 4.25\\
		60 &  4 & 100 & 30.0M & 71.8 & 4.03\\
		60 &  3 & 96 & 28.6M & 70.6 & 3.69\\
		\midrule
		\multicolumn{6}{c}{Varying the \# of queries}\\
		\midrule
		50 &  6 & 104 & 32.8M & 72.6 & 4.48\\
		40 &  6 & 100 & 32.8M & 72.4 & 4.29\\
		30 &  6 & 96 & 32.8M & 71.9 & 4.05 \\
		25 &  6 & 95 & 32.8M & 71.5 & 3.96\\
		\midrule
		\multicolumn{6}{c}{Varying the \# of decoder layers and queries}\\
		\midrule
		25 &  5 & 93 & 31.4M & 71.4 & 3.73\\
		25 &  4 & 91 & 30.0M & 70.7 & 3.53\\
		\bottomrule
	\end{tabular}
\end{table}
\section{Conclusion}
In this paper, we have presented a new real-time end-to-end pose estimator.  DETRPose achieves new state-of-the-art results and faster convergence. Results on multiple datasets demonstrate that DETRPose's excellent performance, proving that DETR-based models can be effectively adapted for pose estimation, opening a promising future for DETR models in the MPPE task. 
A limitation of this work is that DETRPose has higher latency than YOLO model, despite having similar FLOPs. The main reason for this is the group attention, which requires multiple transpose operations, breaking the contiguity of tensors. 

\section*{Acknowledgements}
This work was supported in part by the National Science Foundation under Grant 1949230, Grant1842220, and Grant 1613637. We would like to thank Lambda.ai and UNM Center for Advanced Research Computing for providing the high performance computing resources used in this work.

\newpage
\bibliographystyle{splncs04}
\bibliography{main}

\definecolor{codegreen}{rgb}{0,0.6,0}
\definecolor{codegray}{rgb}{0.5,0.5,0.5}
\definecolor{codepurple}{rgb}{0.58,0,0.82}
\definecolor{backcolour}{rgb}{0.96,0.96,0.96} 

\lstdefinestyle{pythonstyle}{
	backgroundcolor=\color{backcolour},   
	commentstyle=\color{codegray},
	keywordstyle=\color{codepurple}\bfseries,
	numberstyle=\tiny\color{codegray},
	stringstyle=\color{codegreen},
	basicstyle=\ttfamily\footnotesize, 
	breakatwhitespace=false,         
	breaklines=true,                 
	captionpos=b,                    
	keepspaces=true,                 
	numbers=left,                    
	numbersep=5pt,                  
	showspaces=false,                
	showstringspaces=false,
	showtabs=false,                  
	tabsize=2,
	frame=single,                    
	rulecolor=\color{black!20}       
}

\lstset{style=pythonstyle}

\newpage
\appendix

\section{Clarification about RTMO}
We note in Sec. 4.3 that RTMO uses transfer learning during training. This is a delicate part that the paper does not mention, but they used their official GitHub configuration. To avoid breaking any official submission, we do not provide any link, but we can provide steps to verify it. In the mmpose GitHub repository, navigate to  \textit{<mmpose\_repo\_path>/configs/body\_2d\_keypoint/rtmo/coco} and open any Python file. The model variable’s backbone.init cfg dictionary (lines 226-227 for RTMO-L) shows type=’Pretrained’ with a checkpoint parameter specifying pretrained weights. We provide an example of the configuration in \cref{code:rtmo}.

\begin{lstlisting}[float=!h, language=Python, caption=Example for RTMO configuration file., label=code:rtmo]
model = dict(
	type='BottomupPoseEstimator',
	init_cfg= ...,
	data_preprocessor= ...,
	backbone=dict(
		type='CSPDarknet',
		deepen_factor=deepen_factor,
		widen_factor=widen_factor,
		out_indices=(2, 3, 4),
		spp_kernal_sizes=(5, 9, 13),
		norm_cfg=dict(type='BN', momentum=0.03, eps=0.001),
		act_cfg=dict(type='Swish'),
		init_cfg=dict(
			type='Pretrained',
			checkpoint=<checkpoint_path>, # in the repo, you will find a link to a pth file
			prefix='backbone.',
		)
	),
)
\end{lstlisting}

\begin{table}[!t]
	\centering
	\caption{Effects of NMS on AP and latency (ms) for YOLO detection and pose estimation models.}
	\begin{tabular}{l | ccccc}
		\toprule
		& \multicolumn{2}{c}{conf=0.001 }& \phantom{a} & \multicolumn{2}{c}{conf=0.25}\\
		\cmidrule{2-3} \cmidrule{5-6}
		model & AP & ms & & AP & ms \\
		\midrule
		YOLOv8-Detect & 50.0 &  4.26 & & 45.5 & 4.12 ($\downarrow 3.3\%$)\\
		YOLOv8-Pose &  67.3 & 4.13 & & 65.7 & 4.02 ($\downarrow 2.7\%$)\\
		YOLO11-Detect &  50.3 & 3.83 & & 45.7 & 3.68 ($\downarrow 3.9\%$)\\
		YOLO11-Pose &  65.6 & 3.99& & 63.8 &3.86 ($\downarrow 3.3\%$)\\
		\bottomrule
	\end{tabular}
	\label{tab:nms}
\end{table}

\begin{figure}[!t]
	\begin{tabular}{ccc}
		\includegraphics[width=0.3\columnwidth]{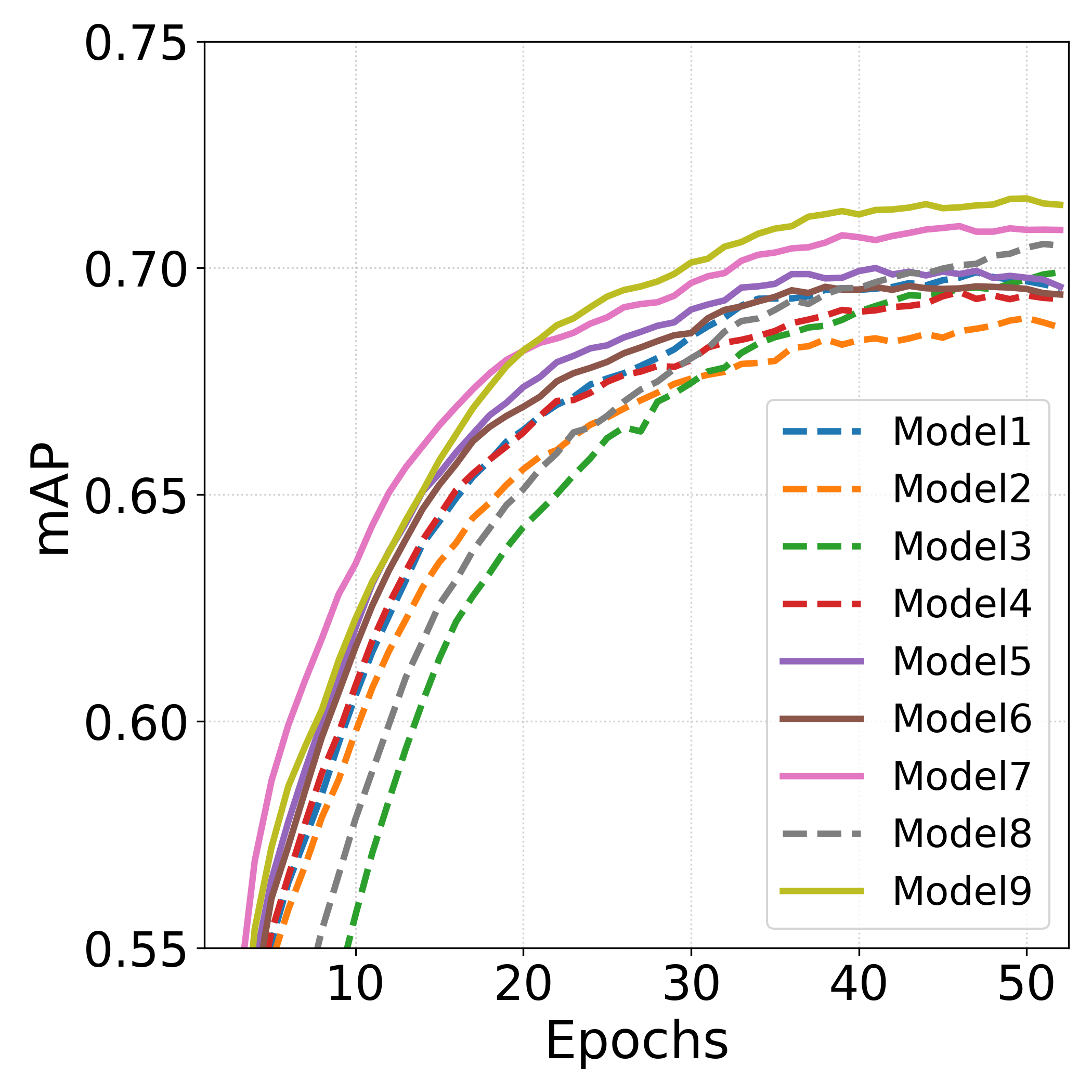} & 
		\includegraphics[width=0.3\columnwidth]{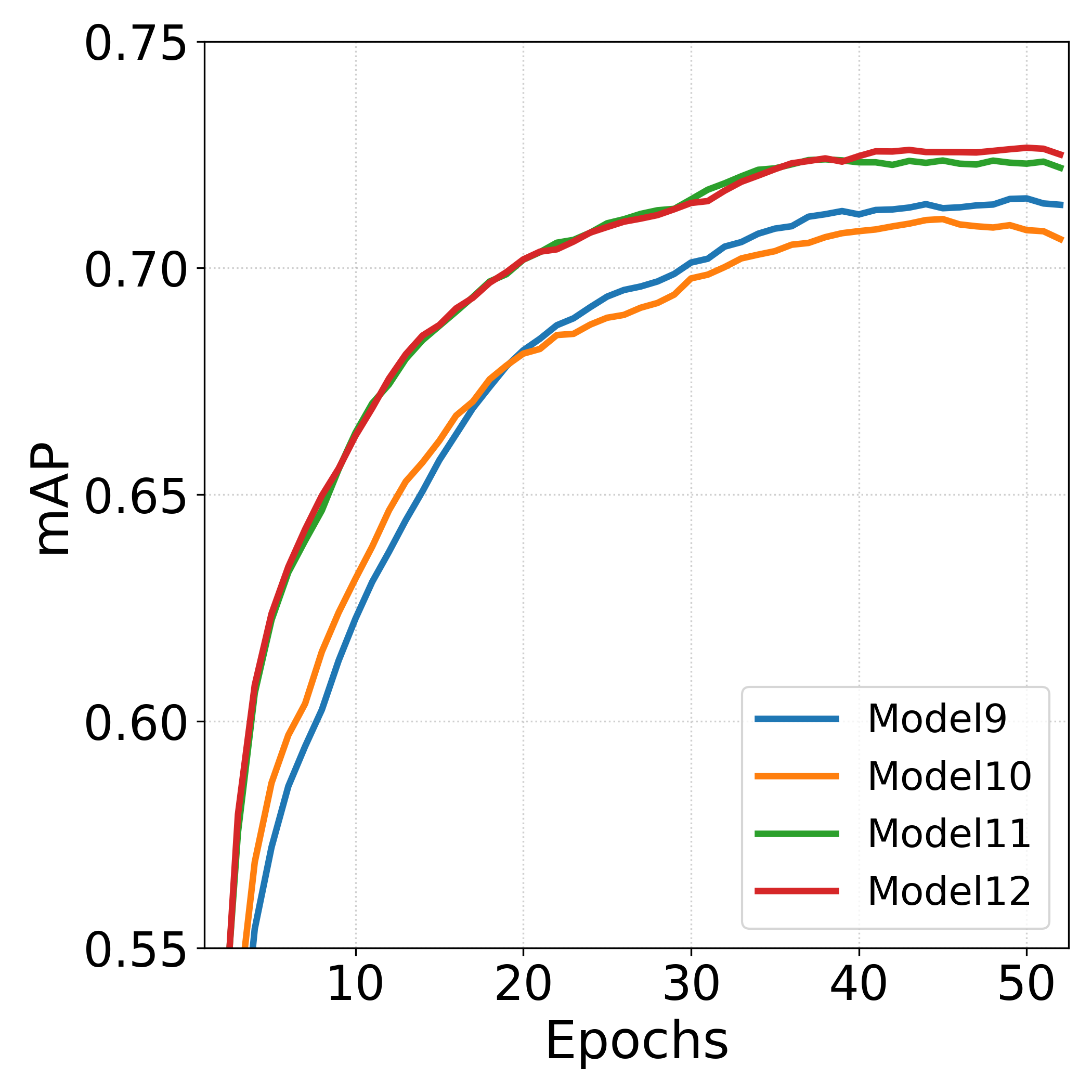} & 
		\includegraphics[width=0.3\columnwidth]{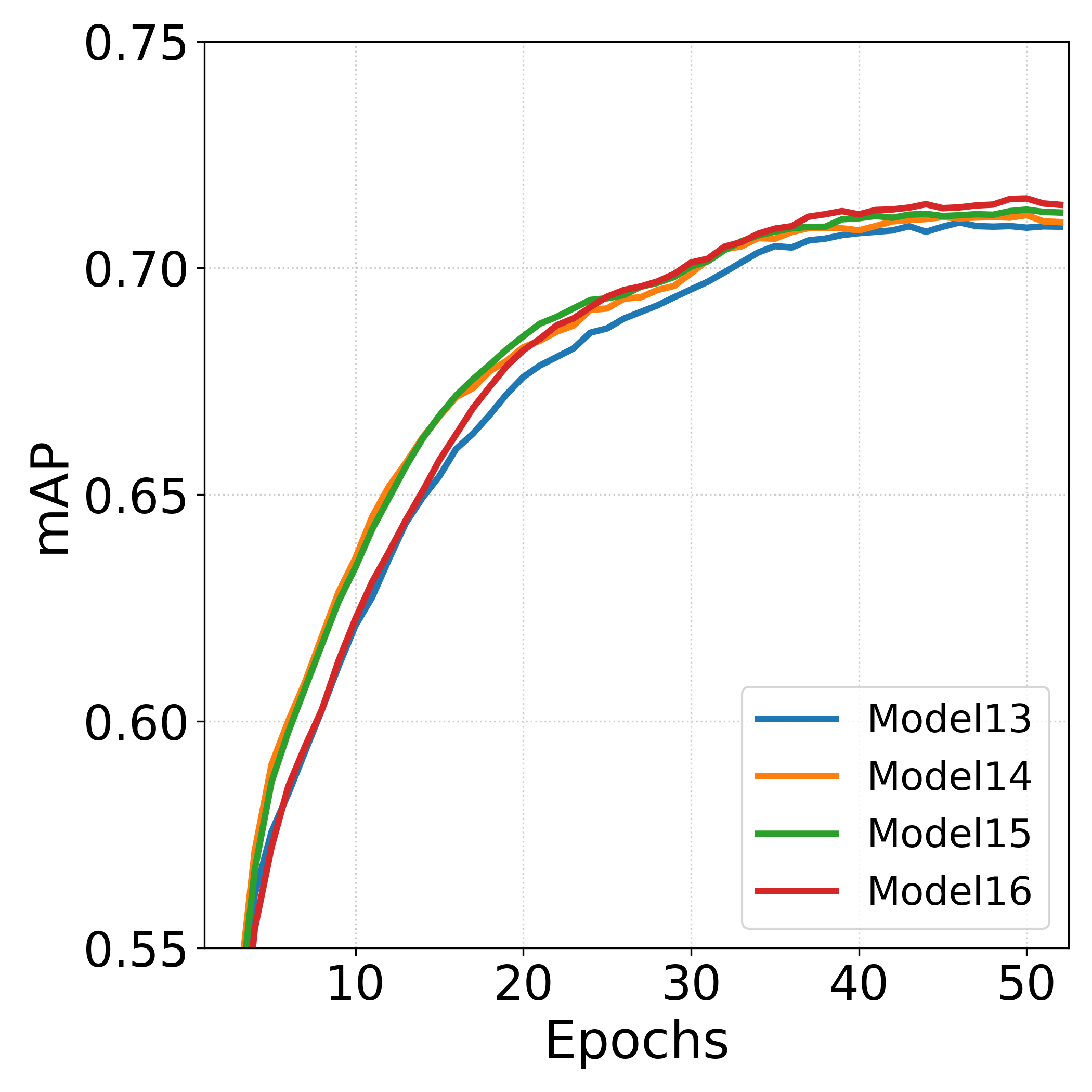}  \\
		a) Architecture & b) Transfer learning & c) Denoising 
	\end{tabular}
	\caption{Convergence analysis for DETRPose-L. Dashed lines represent models trained without the proposed denoising keypoint technique. For information about the names, see Tab. 6 and Tab. 7.}
	\label{fig:appendix_1}
\end{figure}

\section{Effects of NMS in pose estimation models}
Inconsistent latency of Non-Maximum-Surpression (NMS) is an important motivation for real-time end-to-end methods. Although previous works have shown the impact of NMS on the detection task \cite{rtdetr, lwdetr}, its impact on pose estimation requires a separate evaluation. Given that pose estimation tasks typically involve fewer detected instances than object detection, the computational overhead of NMS is expected to be less pronounced. In this study, we systematically evaluate the impact of NMS on both YOLO-based detection and pose estimation models, with quantitative results summarized in \cref{tab:nms}. Our analysis demonstrates that NMS imposes comparable latency constraints across both domains, thereby reinforcing the rationale for employing end-to-end models that ensure consistent inference latency.

\section{Training convergence}
In this section, we evaluate how the components shown in Tab. 6 and Tab. 7 of the main paper affect the convergence of DETRPose-L. \Cref{fig:appendix_1}a shows the effects of each component on the mAP score and the convergence.  Model 1 uses 100 instance queries for detection, while the others use only 60 (for more information, see Tab. 6). We observe that using the proposed KSVF loss or the Pose-LQE layer (models 3 and 4) leads to a better performance, but has a similar or lower convergence rate than model 2. We find it interesting that, with both KSVF and Pose-LQE (model 8), the second-slowest convergence rate is achieved, yet it yields the third-best performance. When we apply the denosing technique (solid lines), the convergence of these models improves. We observe that using Pose-LQE (models 3, 6, 8, 9) negatively affects convergence during the first half of training epochs, but yields better overall results. We believe this occurs because the mosaic augmentation uses letterbox resizing, as in YOLO models \cite{yolo11, rtmo}, to preserve the image aspect ratio, whereas DETR models use square resizing, which changes it. 

\Cref{fig:appendix_1}b shows the effects of transfer learning using D-FINE's pretrained weights (applied only to the backbone and encoder) on convergence. Interestingly, model 9 (without transfer learning) shows a similar convergence to model 10 (COCO-pretrained weights) during the first 20 epochs. Then model 9 achieves a better score. Similarly, model 12 (Object365-pretrained weights) has better performance than model 11 (Object365-to-COCO pretrained weights). As expected, pretraining on the Object365 \cite{objects365} dataset leads to faster convergence in the early epochs, but this gap narrows over time. At the end of training, the difference in mAP between the model without transfer learning and the model initialized with Object365-pretrained weights is 1.2 points, whereas earlier in training, it was greater than 4 points.

\Cref{fig:appendix_1}c shows how the per-keypoint fall-off parameters and the segmented area value affect convergence. We observe that using either only segmented area values (model 14) or a known fall-off parameter (model 15) converges slightly faster than using both (model 16) during the first 20 epochs. Nevertheless, model 16 has slightly better results at the end of the training.

\section{Flexibility of DETRPose}

\begin{figure}[!t]
	\centering
	\begin{tabular}{ccc}
		\includegraphics[width=0.32\columnwidth]{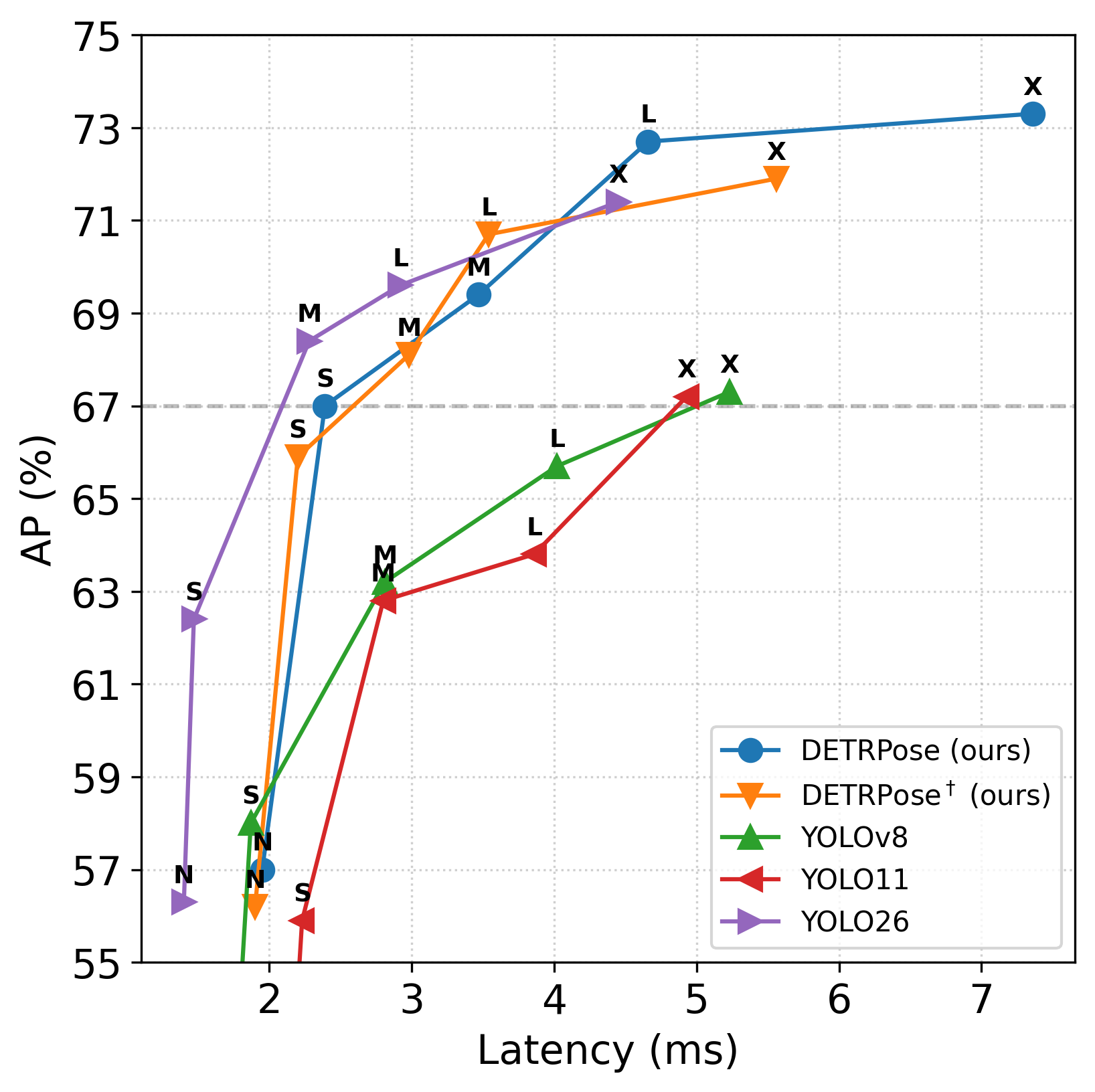} & 
		\includegraphics[width=0.32\columnwidth]{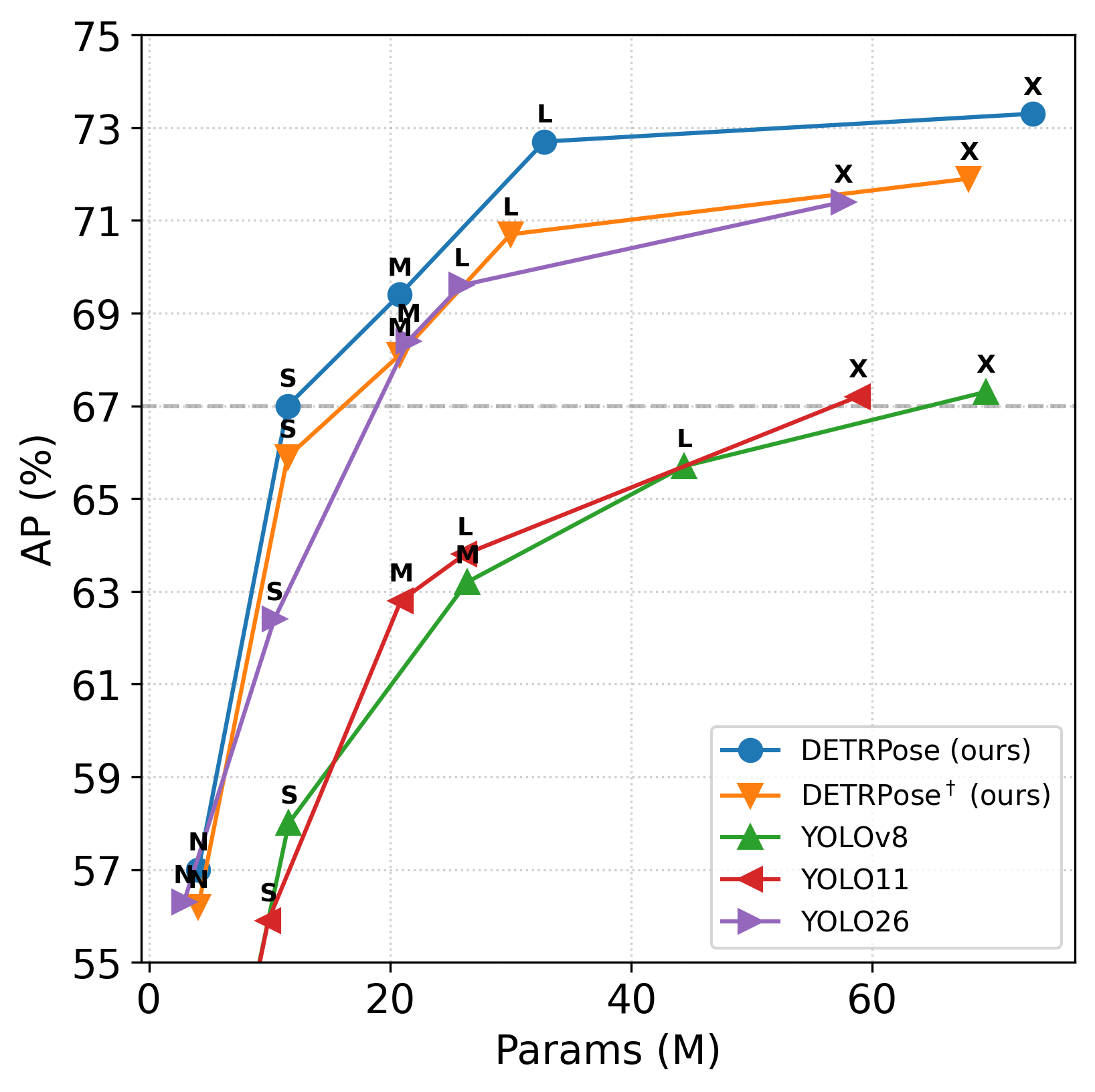} & \includegraphics[width=0.32\columnwidth]{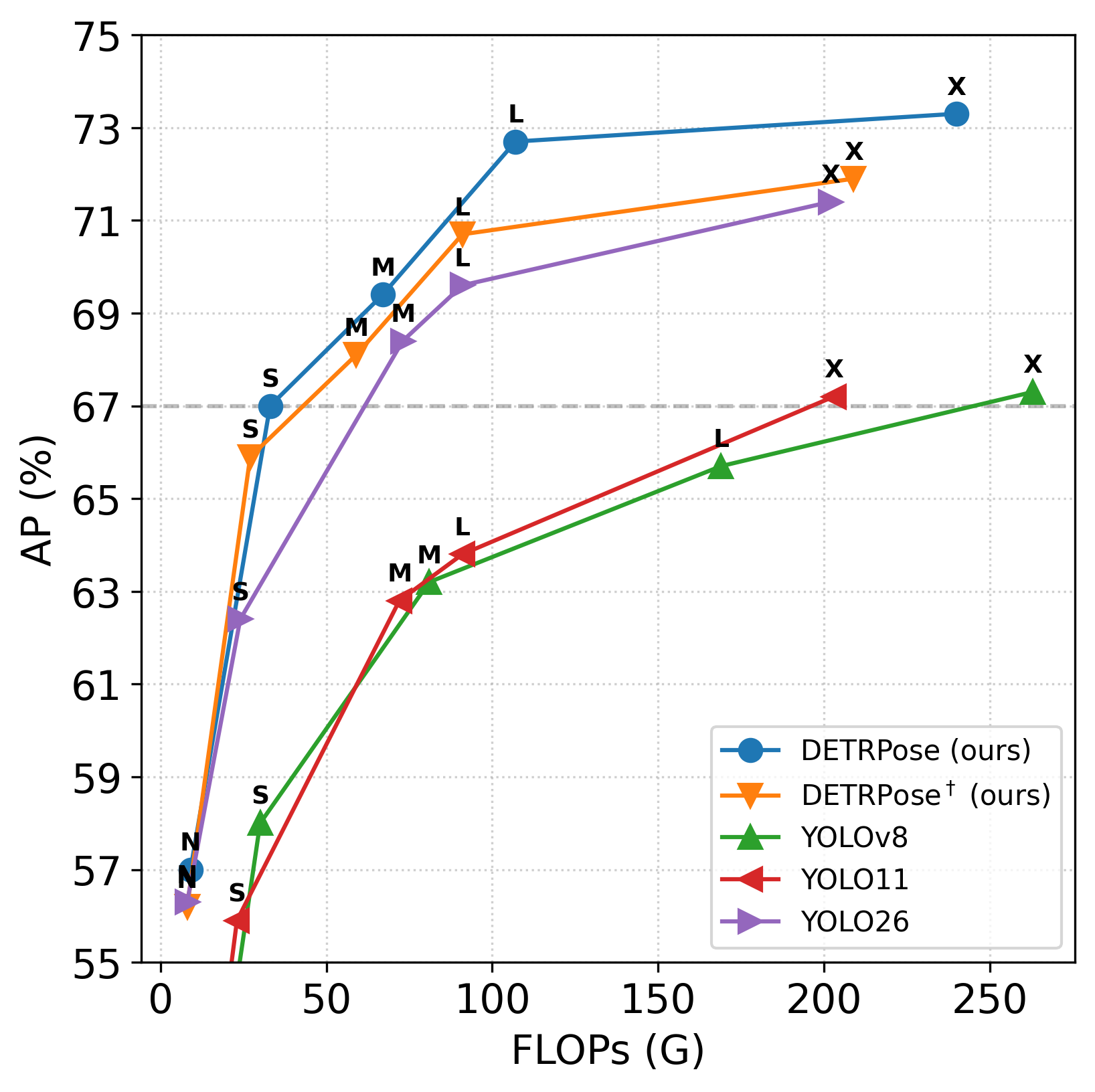}  \\
	\end{tabular}
	\caption{Comparisons with Ultralytics YOLO models in terms of latency (left), model size (mid), and computational cost (right) on the COCO \texttt{val} dataset. DETRPose$^\dagger$ is a modified version of DETRPose with fewer queries or fewer decoder layers. We measure end-to-end latency using TensorRT FP16 on an NVIDIA RTX A5500 GPU.}
	\label{fig:flexi_detrpose}
\end{figure}

\begin{table}[!t]
	\centering
	\caption{Modifications for DETRPose$^\dagger$. The only changes are made to the number of queries and the decoder layers. Values with a $*$ mean that they are the same as in the original DETRPose.}
	\label{tab:modifications}
	\begin{tabular}{c c c c c c}
		\toprule
		Model size & \phantom{a}N\phantom{a} & S & \phantom{a}M\phantom{a} & L & \phantom{a}X\phantom{a}\\
		\midrule
		\# queries & 30 & 25 & 25 & 25 & 25\\
		\# decoder layers & 3$^*$ & 3$^*$ & 4$^*$ & 4 & 4\\
		\bottomrule
	\end{tabular}
\end{table}

\begin{table}[!t]
	\centering\caption{\label{tab:flexi_detrpose}Performance comparison of state-of-the-art one-stage methods on the COCO \texttt{val} dataset. The best and second results are in \textbf{bold} and \textit{italic}, respectively. DETRPose$^\dagger$ is a modified version of DETRPose with fewer queries or fewer decoder layers. Latency is reported using an NVIDIA RTX A5500 GPU.}
	\resizebox{\columnwidth}{!}{
		\begin{tabular}{l c c ccc ccc}
			\toprule 
			Method & \#Params & \phantom{a}Time (ms)\phantom{a} & GFLOPs & \phantom{a}AP\phantom{a} & $\textrm{AP}_{50}$ & \phantom{a}$\textrm{AP}_{75}$\phantom{a} & $\textrm{AP}_{M}$ & \phantom{a}$\textrm{AP}_{L}$\phantom{a} \\
			\midrule 
			\multicolumn{9}{c}{Non-end-to-end methods}\\
			\midrule
			YOLOv8-N \cite{yolov8} & 3.3M & 1.67 & 9 & 47.8 & 73.9 & 51.9 & 37.2 & 62.2\\
			YOLOv8-S \cite{yolov8} & 11.6M & 1.87 & 30 & 58.0 & 82.4 & 64.7 & 49.8 & 70.0 \\
			YOLOv8-M \cite{yolov8} & 26.4M & 2.81 & 81 & 63.2 & 85.6 & 70.8 & 56.6 & 73.1 \\
			YOLOv8-L \cite{yolov8} & 44.4M & 4.02 & 169 & 65.7 & 86.6 & 73.0 & 59.4 & 75.1 \\
			YOLOv8-X \cite{yolov8} & 69.4M & 5.23 & 263 & 67.3 & 87.5 & 75.0 & 61.5 & 76.1 \\
			YOLO11-N \cite{yolo11} & 2.9M & 2.05 & 7 & 47.0 & 74.8 & 51.0 & 37.6 & 60.5 \\
			YOLO11-S \cite{yolo11} & 9.9M & 2.23 & 23 & 55.9 & 81.2 & 62.1 & 47.8 & 67.6 \\
			YOLO11-M \cite{yolo11} & 20.9M & 2.80 & 72 & 62.8 & 85.4 & 70.6 & 55.9 & 71.4 \\
			YOLO11-L \cite{yolo11} & 26.2M & 3.86 & 91 & 63.8 & 86.2 & 71.4 & 58.2 & 72.5 \\
			YOLO11-X \cite{yolo11} & 58.8M & 4.93 & 203 & 67.2 & 87.5 & 74.8 & 62.3 & 75.5 \\
			YOLO26-N \cite{yolo11} & 2.9M & 1.40 & 8 & 56.3 & 81.8 & 61.3 & 47.7 & 69.0 \\
			YOLO26-S \cite{yolo11} & 10.4M & 1.47 & 24 & 62.4 & 85.4 & 68.4 & 54.5 & 73.9  \\
			YOLO26-M \cite{yolo11} & 21.5M & 2.28 & 73 & 68.4 & 89.2 & 75.3 & 62.4 & 77.9 \\
			YOLO26-L \cite{yolo11} & 25.9M & 2.92 & 91 & 69.6 & 89.6 & 76.3 & 64.1 & 78.7 \\
			YOLO26-X \cite{yolo11} & 57.6M & 4.45 & 202 & 71.4 & 90.9 & 78.5 & 65.9 & 80.0 \\
			\midrule
			\multicolumn{9}{c}{End-to-end methods}\\
			\midrule
			DETRPose-N (ours) & 4.1M & 1.55 & 9 & 57.0 & 82.2 & 60.8 & 48.0 & 70.3\\
			DETRPose-N$^\dagger$(ours) & 4.1M & 1.50 & 8 & 56.2 & 81.1 & 60.1 & 46.8 & 69.9\\
			DETRPose-S (ours) &  11.5M & 2.39 & 33 & 67.0 & 88.1 & 72.9 & 60.4 & 77.3\\
			DETRPose-S$^\dagger$ (ours) & 11.5M & 2.20 & 27 & 56.2 & 81.1 & 60.1 & 54.5 & 73.9\\
			DETRPose-M (ours) & 20.8M & 3.47 & 67 & 69.4 & 89.8 & 75.5 & 63.1 & 79.1 \\
			DETRPose-M$^\dagger$ (ours) & 20.8M & 2.98 & 59 & 68.1 & 88.4 & 74.1 & 61.1 & 78.6 \\
			DETRPose-L (ours) & 32.8M & 4.66 & 107 & 72.7 & 91.0 & 79.2 & 66.7 & 82.2\\
			DETRPose-L$^\dagger$ (ours) & 30.0M & 3.54 & 91 & 70.7 & 89.8 & 77.2 & 64.0 & 80.7\\
			DETRPose-X (ours) & 73.3M & 7.36 & 240 & 73.3 & 90.5 & 79.4 & 67.5 & 82.7\\   
			DETRPose-X$^\dagger$ (ours) & 68.0M & 5.56 & 209 & 71.9 & 89.7 & 78.3 & 65.8 & 81.5\\
			\bottomrule
		\end{tabular}
	}
\end{table}

Recently, YOLO26 was released by Ultralytics \cite{yolo26}, the second YOLO model to remove NMS post-processing. Since YOLO26 is NMS-free, it achieves lower latency than its predecessors. Our proposed DETRPose model is already slower than YOLOv8 \cite{yolov8} and YOLO11 \cite{yolo11}. To make DETRPose competitive with Ultralytics YOLO models, we reduce the number of queries and decoder layers (see \cref{tab:modifications}). These modifications do not require retraining, since each decoder layer was trained independently. We present visual comparisons in \cref{fig:flexi_detrpose} and qualitative results in \cref{tab:flexi_detrpose}.

We observe that DETRPose-L$^\dagger$ and DETRPose-X$^\dagger$ have a big change in latency. DETRPose-L$^\dagger$ is faster than YOLOv11 and YOLOv8 and still produces better results. DETRPose-X$^\dagger$ shows competitive latency, being 0.33 ms slower than YOLOv8-X. DETRPose$^\dagger$-M is slower by only 0.17 ms compared to YOLOv8-M and YOLO-v11, but achieves better performance (+5.0 points).  We observed that our two smallest models (N and S) exhibit lower latency, but this does not offset the reduced scores. This occurs because the model size is too small and because the number of decoder layers is already 3.

Compared to YOLO26, DETRPose$^\dagger$ has higher latency. We observed that DETRPose$^\dagger$-S has a latency similar to YOLO26-M, but YOLO26-M achieves a higher score. This occurs because YOLO26 uses a better keypoint loss. For DETRPose-L$^\dagger$, it is almost midway between YOLO26-L and YOLO26-X in both latency and AP. DETRPose-X$^\dagger$ is also slower than YOLO26-X but yields better results. This means that DETRPose-X$^\dagger$ can be used as an alternative to YOLO26-X to achieve better results at the cost of 1.1 ms.

\section{OCHuman Dataset Comparisons}
\begin{figure}[!t]
	\centering
	\begin{tabular}{cc}
		\includegraphics[width=0.5\columnwidth]{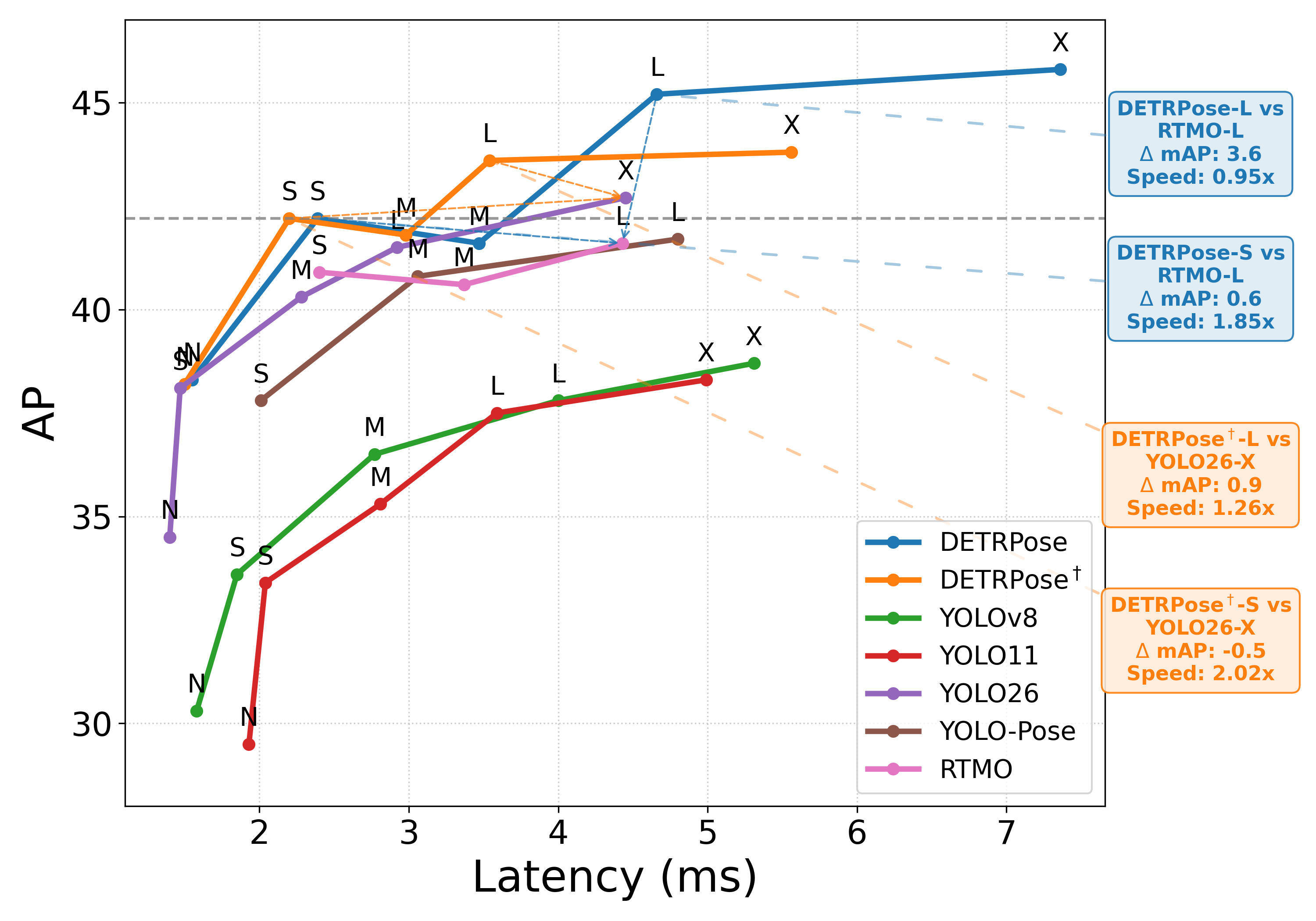} &  \includegraphics[width=0.5\columnwidth]{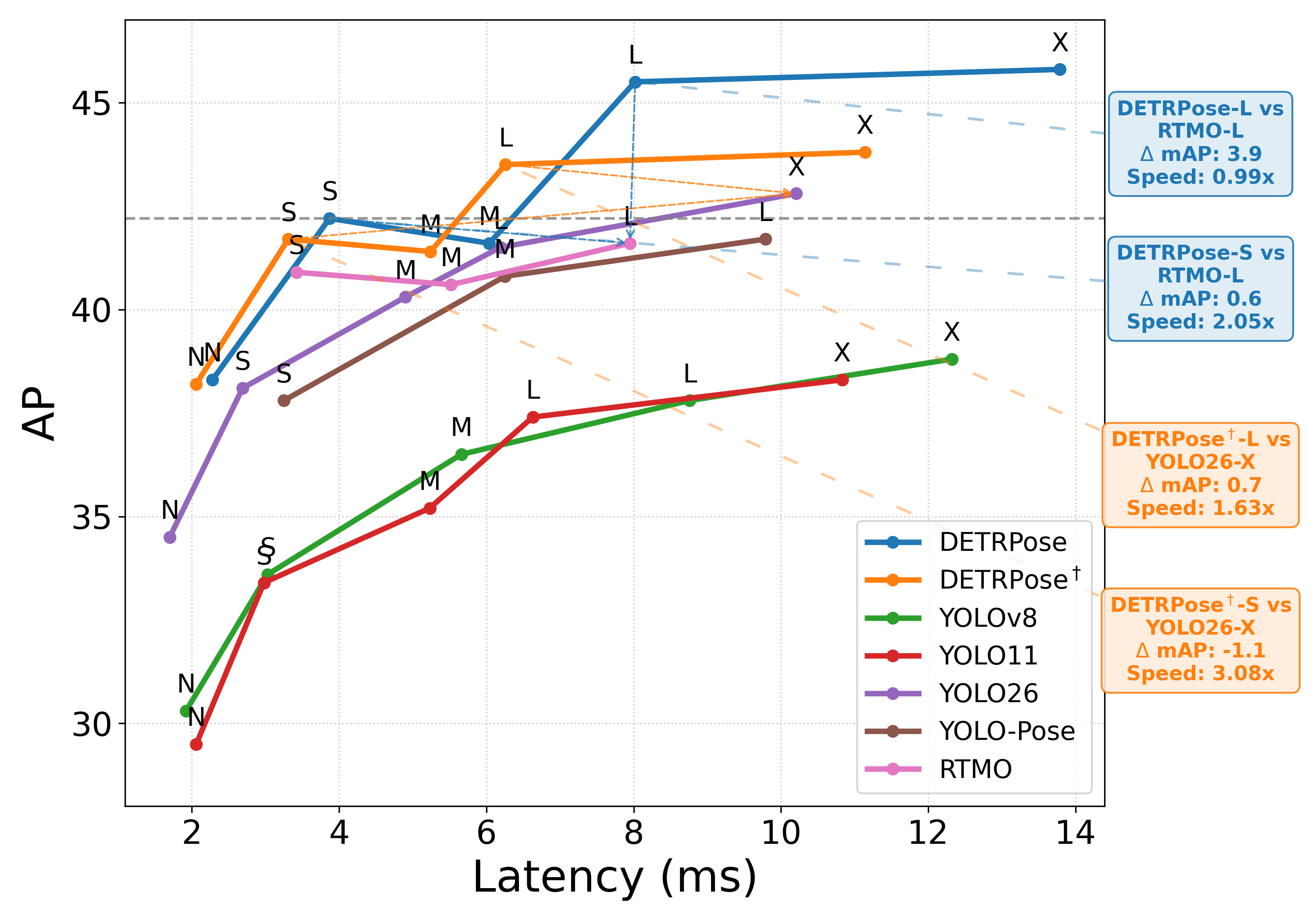}  \\
		a) Floating Point 16  & b) Floating Point 32
	\end{tabular}
	\caption{Comparisons with YOLO models on the OCHuman dataset.  DETRPose$^\dagger$ is a modified version of DETRPose with fewer queries or fewer decoder layers. We measure end-to-end latency using TensorRT on an NVIDIA RTX A5500 GPU.}
	\label{fig:oc_human_sup}
\end{figure}

We evaluate the robustness of the COCO-trained models on the OCHuman dataset and report qualitative results on  \cref{tab:oc_human_sup} and a visual comparison on \cref{fig:oc_human_sup}. YOLO26 significantly improves the AP compared to previous Ultralytics YOLO models. DETRPose-S$^\dagger$ has 0.5 points less in AP than YOLO26-X while being $2\times$ faster. Similarly, DETRPose-L$^\dagger$ improves YOLO26-X by 0.9 and reduces latency by 25\%.  The results demonstrate the strong performance of DETRPose compared with more recent pose estimation models that employ more advanced estimation methods.

\begin{table*}[!t]
	\centering\caption{\label{tab:oc_human_sup}Performance comparison on the OCHuman dataset.  DETRPose$^\dagger$ is a modified version of DETRPose with fewer queries or fewer decoder layers. Best and second results are in \textbf{bold} and \textit{italic}.}
	\begin{tabular}{l cc ccc c}
		\toprule
		Method & \# Params & \phantom{a}Time\phantom{a}  & \phantom{a}AP\phantom{a} & $\textrm{AP}_{50}$ & \phantom{a}$\textrm{AP}_{75}$\phantom{a} & $\textrm{AP}_{L} $\\
		& & (ms)  & & & & \\
		\midrule
		$\textrm{AP}_{L} $\\
		& & (ms) & & & & \\
		\midrule
		YOLO-Pose-S~\cite{yolopose} & 10.8M & 2.01 & 37.8 & 52.6 & 42.9 & 73.4\\
		YOLO-Pose-M~\cite{yolopose}  & 29.3M & 3.06 & 40.8 & 53.2 & 46.1 & 78.2\\
		YOLO-Pose-L~\cite{yolopose} & 61.3M & 4.80 & 41.7 & 53.2 & 47.0 & 79.4\\
		RTMO-S \cite{yolo11} & 9.9M & 2.40 & 40.5 & 52.8 & 45.4 & 77.3\\
		RTMO-M \cite{yolo11} & 22.6M & 3.37 & 39.9 & 50.6 & 44.6 & 79.7\\
		RTMO-L \cite{yolo11} & 44.7M & 4.43 & 40.6 & 50.5 & 45.1 & 81.0\\
		YOLOv8-N \cite{yolov8} & 3.3M & 1.59 & 30.3 & 48.4 & 32.5 & 61.5\\
		YOLOv8-S \cite{yolov8} & 11.6M & 1.86 & 33.6 & 49.3 & 37.4 & 68.3\\
		YOLOv8-M \cite{yolov8} & 26.4M & 2.79 & 36.5 & 50.3 & 40.8 & 72.5\\
		YOLOv8-L \cite{yolov8} & 44.4M & 4.02 & 37.8 & 51.1 & 42.4 & 74.4\\
		YOLOv8-X \cite{yolov8} & 69.4M & 5.30 & 38.7 & 51.1 & 43.3 & 76.3\\
		YOLO11-N\cite{yolo11} & 2.9M & 1.93 & 29.5 & 48.8 & 31.6 & 60.5\\
		YOLO11-S \cite{yolo11} & 9.9M & 2.05 & 33.4 & 49.6 & 37.6 & 67.4\\
		YOLO11-M \cite{yolo11} & 20.9M & 2.82 & 35.3 & 49.7 & 39.9 & 71.5\\
		YOLO11-L \cite{yolo11} & 26.2M & 3.61 & 37.5 & 51.4 & 42.1 & 73.0\\
		YOLO11-X \cite{yolo11} & 58.8M & 5.01 & 38.3 & 51.1 & 43.2 & 75.0\\
		YOLO26-N \cite{yolo11} & 2.9M & 1.25 & 34.5 & 51.6 & 38.4 & 72.3\\
		YOLO26-S \cite{yolo11} & 10.4M & 1.42 & 38.1 & 53.2 & 42.8 & 77.0\\
		YOLO26-M \cite{yolo11} & 21.5M & 2.28 & 40.3 & 53.2 & 44.9 & 81.7\\
		YOLO26-L \cite{yolo11} & 25.9M & 2.91 & 41.5 & 53.9 & 46.5 & 82.7\\
		YOLO26-X \cite{yolo11} & 57.6M & 4.45 & 42.7 & 54.2 & 47.6 & 84.4\\
		\midrule
		\multicolumn{7}{c}{End-to-end methods}\\
		\midrule
		DETRPose-N (ours) & 4.1M & 1.61 & 38.3 & 53.7 & 43.2 & 80.0\\
		DETRPose-N$^\dagger$ (ours) & 4.1M & 1.66 & 38.2 & 53.6 & 43.0 & 77.5\\
		DETRPose-S (ours) & 11.5M & 2.41 & 42.2 & 54.3 & 47.5 & 85.2\\
		DETRPose-S$^\dagger$ (ours) & 11.5M & 2.00 & 42.2 & 54.6 & 47.3 & 81.9\\
		DETRPose-M (ours) & 20.8M & 3.59 & 41.6 & 53.0 & 46.9 & 86.2\\
		DETRPose-M$^\dagger$ (ours) & 20.8M & 2.98 & 41.8 & 53.6 & 47.1 & 83.4\\
		DETRPose-L (ours) & 32.8M & 4.83 & \textit{45.2} & \textit{55.4} & \textit{50.4} & \textbf{89.0}\\
		DETRPose-L$^\dagger$ (ours) & 30.0M & 3.55 & 43.6 & 54.5 & 49.0 & 85.3\\
		DETRPose-X (ours) & 73.3M & 7.41 & \textbf{45.8} & \textbf{55.7} & \textbf{51.1} & \textit{87.8}\\
		DETRPose-X$^\dagger$ (ours) & 68.0M & 5.54 & 43.8 & 54.5 & 49.5 & 84.5\\
		\bottomrule
	\end{tabular}
\end{table*}

\end{document}